\documentclass[10pt,twocolumn,letterpaper]{article}

\usepackage{cvpr}              % To produce the CAMERA-READY version
\usepackage{graphicx}
\usepackage{amsmath}
\usepackage{amssymb}
\usepackage{amsfonts}   % 直接提供 \mathbb
\usepackage{booktabs}
\usepackage{multirow}
\usepackage[table]{xcolor}
\usepackage{enumitem}
\usepackage{pifont}
\usepackage{makecell}   % \thead
\usepackage{array}      % 列格式增强
\usepackage{algorithm}
\usepackage{capt-of} % 提供 \captionof{figure}
\usepackage[noend]{algpseudocode}
\usepackage{subcaption}   % provides subtable
\usepackage{wasysym}
\usepackage[most]{tcolorbox}
\usepackage{amsthm}

\theoremstyle{plain}
\theoremstyle{definition}

\theoremstyle{remark}

\definecolor{cvprblue}{rgb}{0.21,0.49,0.74}
\usepackage[pagebackref,breaklinks,colorlinks,allcolors=cvprblue]{hyperref}

\usepackage[capitalize]{cleveref}
\crefname{section}{Sec.}{Secs.}
\Crefname{section}{Section}{Sections}
\Crefname{table}{Table}{Tables}
\crefname{table}{Tab.}{Tabs.}

\definecolor{upcolor}{RGB}{57,182,74}
\definecolor{local}{RGB}{78,149,217}
\definecolor{global}{RGB}{61,143,115}
\definecolor{myred}{RGB}{192,0,0}
\definecolor{mygreen}{RGB}{71,212,90}
\definecolor{pur}{RGB}{213,205,239}
\definecolor{blu}{RGB}{210,228,245}
\definecolor{SoftLavender}{HTML}{F0E9F3}
\definecolor{LavenderEdge}{HTML}{BBBADA}

\newcommand{\up}[1]{\textcolor{upcolor}{\textcolor{upcolor}{$\uparrow$}#1}}

\newcommand{\cmark}{\textcolor{green!60!black}{\ding{51}}} % ✓
\newcommand{\xmark}{\textcolor{red!70!black}{\ding{55}}}   % ✗
\DeclareMathOperator*{\argmax}{arg\,max}

\newenvironment{tightEq}[1][6pt]{%
  \begingroup
  \setlength{\abovedisplayskip}{#1}%
  \setlength{\belowdisplayskip}{#1}%
  \setlength{\abovedisplayshortskip}{#1}%
  \setlength{\belowdisplayshortskip}{#1}%
  \ignorespaces
}{%
  \endgroup\ignorespacesafterend
}

\newtcolorbox{purpleprompt}[2][]{%
  enhanced,
  colback=cvprblue!5,
  colframe=cvprblue,
  title={#2},
  float,                 % 使其成为“浮动体”
  floatplacement=t,      % 默认置顶；实例可用 [floatplacement=!t]
  listing above text,
}

\def\paperID{} % *** Enter the Paper ID here
\def\confName{CVPR}
\def\confYear{2026}

\title{DeepScan: A Training-Free Framework for Visually Grounded Reasoning \\ in Large Vision-Language Models}

\author{%
  \hspace*{-2mm}%
  \makebox[\textwidth][c]{%
    {\setlength{\tabcolsep}{5pt}% 
    \begin{tabular}{cccccc}
      {\small Yangfu Li$^{\,\clubsuit}$} & {\small Hongjian Zhan$^{\,\clubsuit,}$}\thanks{Corresponding Author. \emph{This work is accepted to CVPR'26.}} 
      & {\small Jiawei Chen$^{\,\clubsuit}$} & {\small Yuning Gong$^{\,\heartsuit,\,\diamondsuit}$} & {\small Qi Liu$^{\,\clubsuit}$} & {\small Yue Lu$^{\,\clubsuit}$}
    \end{tabular}%
    }%
  }\\ %[1ex]
  \multicolumn{1}{c}{\small\normalfont $^{\clubsuit}$East China Normal University,\quad $^{\heartsuit}$Sichuan University,\quad $^{\diamondsuit}$Shanghai AI Laboratory}\\
  \multicolumn{1}{c}{\small\texttt{\{yfli\_cee, qiliu\}@stu.ecnu.edu.cn, hjzhan@cee.ecnu.edu.cn, ylu@cs.ecnu.edu.cn}} \\
}

\makeatletter
\g@addto@macro\@maketitle{%
  \vspace{-9mm}
  \begin{center}
    \includegraphics[width=0.95\textwidth]{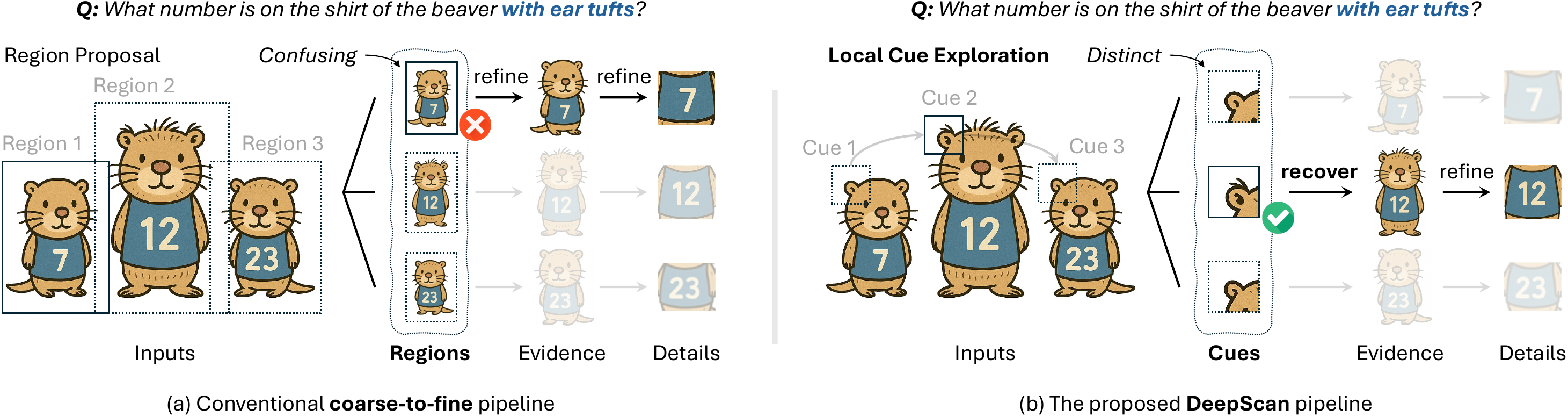}
    \vspace{-1mm}
    \captionof{figure}{Comparison of grounding pipelines in existing methods and the proposed DeepScan. (a) Existing methods depend on \emph{one-shot} localization of evidence regions, making them sensitive to noisy contexts. (b) DeepScan uses local cue exploration to identify discriminative cues and recovers evidence from these cues in a \emph{bottom-up} manner, robustly localizing critical visual content even in challenging scenes.
    }
    \label{fig:1}
  \end{center}
}
\def\thm@space@setup{%
  \thm@preskip=2pt  
  \thm@postskip=2pt  
}
\renewenvironment{abstract}{%
  \centerline{\large\bf Abstract}%
  \vspace{5pt}% 
  \it
}{\par}
\makeatother

\begin{document}
\maketitle

\begin{abstract}
Humans can robustly localize visual evidence and provide grounded answers even in noisy environments by identifying critical cues and then relating them to the full context in a bottom-up manner. Inspired by this, we propose DeepScan, a training-free framework that combines Hierarchical Scanning, Refocusing, and Evidence-Enhanced Reasoning for visually grounded reasoning in Large Vision–Language Models (LVLMs). Unlike existing methods that pursue one-shot localization of complete evidence, Hierarchical Scanning performs local cue exploration and multi-scale evidence extraction to recover evidence in a bottom-up manner, effectively mitigating the impacts of distractive context. Refocusing then optimizes the localized evidence view through collaboration of LVLMs and visual experts. Finally, Evidence-Enhanced Reasoning aggregates multi-granular views via a hybrid evidence memory and yields accurate and interpretable answers. Experimental results demonstrate that DeepScan significantly boosts LVLMs in diverse visual tasks, especially in fine-grained visual understanding. It achieves 90.6\% overall accuracy on V* when integrated with Qwen2.5-VL-7B. Moreover, DeepScan provides consistent improvements for LVLMs across various architectures and model scales without additional adaptation cost. The code will be open-source at \href{https://github.com/YChenL/DeepScan}{\textcolor{cvprblue}{\textbf{DeepScan}}}.
\end{abstract}    
\section{Introduction}

Humans excel at grounding critical visual information and reasoning based on the located evidence, which is a hallmark of intelligence recognized in the cognitive and vision science.~\cite{wang2023statistical,wolfe2020visual,wolfe2017five}. While this is natural to humans, it remains challenging for Large Vision-Language Models (LVLMs) to replicate this behavior, leading to suboptimal performance in complex visual tasks~\cite{wangscaling,wu2024vstar,wang2025hrbench,li20251+}.

To reproduce this ability in LVLMs, some approaches have employed reinforcement learning~\cite{su2025pixelreasoner,zheng2025deepeyes,wang2025traceable,zhang2025thyme} with visually supervised rewards (\eg, IoU) or context engineering~\cite{wu2024vstar,yu2025zoom}, enabling active perception and localization of evidence during reasoning. Another line of research augments LVLM grounding with auxiliary modules~\cite{chen2023shikra,Rasheed2024GLaMM,You2024Ferret,Zhang2025PSALM}. More recently, several studies~\cite{li2025dyfo,an2025mitigating,qian2025zoomer} integrate external visual experts, such as GroundingDINO~\cite{liu2024grounding,ren2024grounding} and LangSAM~\cite{medeiros2024language}, to localize visual evidence based on the consensus between these experts and LVLMs, significantly enhancing fine-grained visual understanding in LVLMs.

Despite these advances, most existing methods follow a \emph{top-down} grounding paradigm: they first perform an image-level search for coarse-grained proxies, such as region proposals~\cite{su2025pixelreasoner,zheng2025deepeyes,wang2025traceable,zhang2025thyme,yu2025zoom}, detection boxes~\cite{li2025dyfo,qian2025zoomer}, or textual descriptions~\cite{wang2025hrbench}, and then refine these proxies to obtain fine-grained evidence. However, this paradigm requires the \emph{one-shot} localization of complete evidence regions from the entire image, which is easily affected by noisy context, \ie, attention sink~\cite{xiaoefficient}, or semantically similar objects, \ie, attention drift~\cite{cheng2017focusing}, leading to suboptimal performance. In such cases, the LVLM either refuses to answer or makes uninformed guesses based on incorrect evidence, as illustrated in~\cref{fig:1}. By contrast, humans intuitively adopt a \emph{bottom-up} routine in challenging visual tasks. For example, when playing the ``\emph{spot-the-difference}'' puzzles, people scan local patches for subtle discrepancies within largely similar content, then verify these cues at the image level to recover the target while suppressing distracting context.

\begin{figure}[t]
	\centering
	\includegraphics[width=1\linewidth]{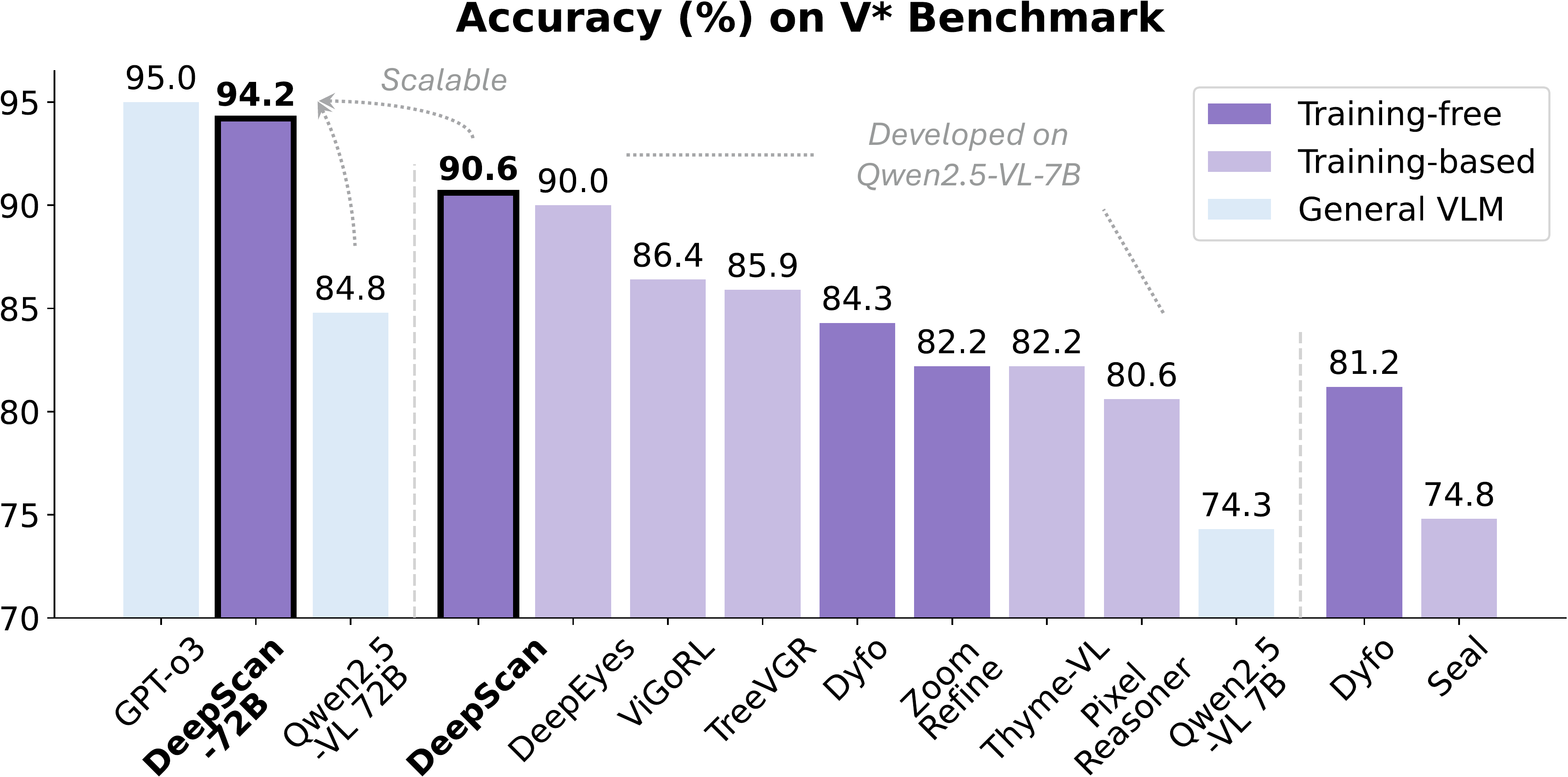}
	\caption{Performance of LVLMs and visually grounded reasoning variants on V*. DeepScan achieves highly competitive results.}
	\label{fig:2}
    \vspace{-1mm}
\end{figure}

Inspired by this human behavior, we propose \textbf{DeepScan}, a training-free framework for visually grounded reasoning in LVLMs, comprising Hierarchical Scanning, Refocusing, and Evidence-Enhanced Reasoning.
Unlike existing methods, Hierarchical Scanning leverages patch-wise cue exploration and image-level evidence extraction to robustly localize visual evidence in a bottom-up manner. Refocusing further refines the surrounding context of the extracted evidence through collaboration between the visual experts and LVLMs, reducing information loss. Finally, Evidence-Enhanced Reasoning utilizes a Hybrid Evidence Memory to provide LVLMs with multi-granular evidence views, producing accurate and well-grounded answers. \cref{fig:2} shows that DeepScan performs competitively with leading approaches. Moreover, as a training-free framework, DeepScan scales seamlessly to larger models (\eg, Qwen2.5-VL-72B), yielding consistently improved performance. Extensive experiments across diverse visual tasks further demonstrate its effectiveness. Our contributions are fourfold:
\begin{enumerate}[leftmargin=*,
                  itemsep=2pt,    
                  parsep=0pt,    
                  topsep=2pt,     
                  partopsep=0pt]  
    \item[\ding{182}] We introduce \textbf{DeepScan}, a training-free framework that boosts LVLM performance by explicitly localizing, recalibrating, and integrating evidence before answering.
    \item[\ding{183}] We propose \textbf{Hierarchical Scanning}, a bottom-up visual grounding paradigm that mitigates noisy context via local cue exploration and multi-scale evidence extraction.
    \item[\ding{184}] We present \textbf{Refocusing}, a collaborative search paradigm that recalibrates the evidence view through interactions between LVLMs and external visual experts.
    \item[\ding{185}] Comprehensive experiments validate the superiority of DeepScan. It provides average improvements of 16.3\% on V* and 5.5\% on TreeBench for Qwen2.5-VL-7B. Detailed ablation studies across LVLM architectures and scales further validate the generalizability of our method.
\end{enumerate}
\section{Related Work}
\subsection{Large Vision-Language Models}
Early breakthroughs in Large Vision-Language Models (LVLMs), such as Flamingo~\cite{alayrac2022flamingo} and BLIP-2~\cite{li2023blip}, integrated visual features into an LLM via cross-attention. In contrast, LLaVA~\cite{liu2023llava} maps features from a frozen vision encoder (\eg, CLIP~\cite{radford2021learning}) into the LLM’s semantic space with a lightweight MLP. This feature-projection paradigm has driven rapid progress, with subsequent work scaling LVLMs and tackling increasingly complex tasks, such as OCR~\cite{zhan2024free,li2025msa2,zhan2024fare}, general VQA~\cite{liu2024llavanext,li2024llavaov,zhu2025internvl3}, and latent visual reasoning~\cite{zhang2025latent,li2024mvot}. A critical frontier is to handle high-resolution input. For instance, LLaVA-NeXT~\cite{liu2024llavanext} and InternVL1.5~\cite{chen2024internvl15} support arbitrary resolutions, while Qwen2-VL~\cite{wang2024qwen2vl} and Qwen2.5-VL~\cite{bai2025qwen25vl} introduce mRoPE for resolution generality. Despite these advances, current LVLMs lack an explicit mechanism to perceive and localize task-relevant visual evidence, which leads to hallucinations, especially in high-resolution scenes. We address this gap with DeepScan, a training-free framework that couples LVLMs with external experts to achieve visually grounded reasoning, yielding more accurate and interpretable answers.

\begin{figure*}[t]
	\centering
	\includegraphics[width=1\linewidth]{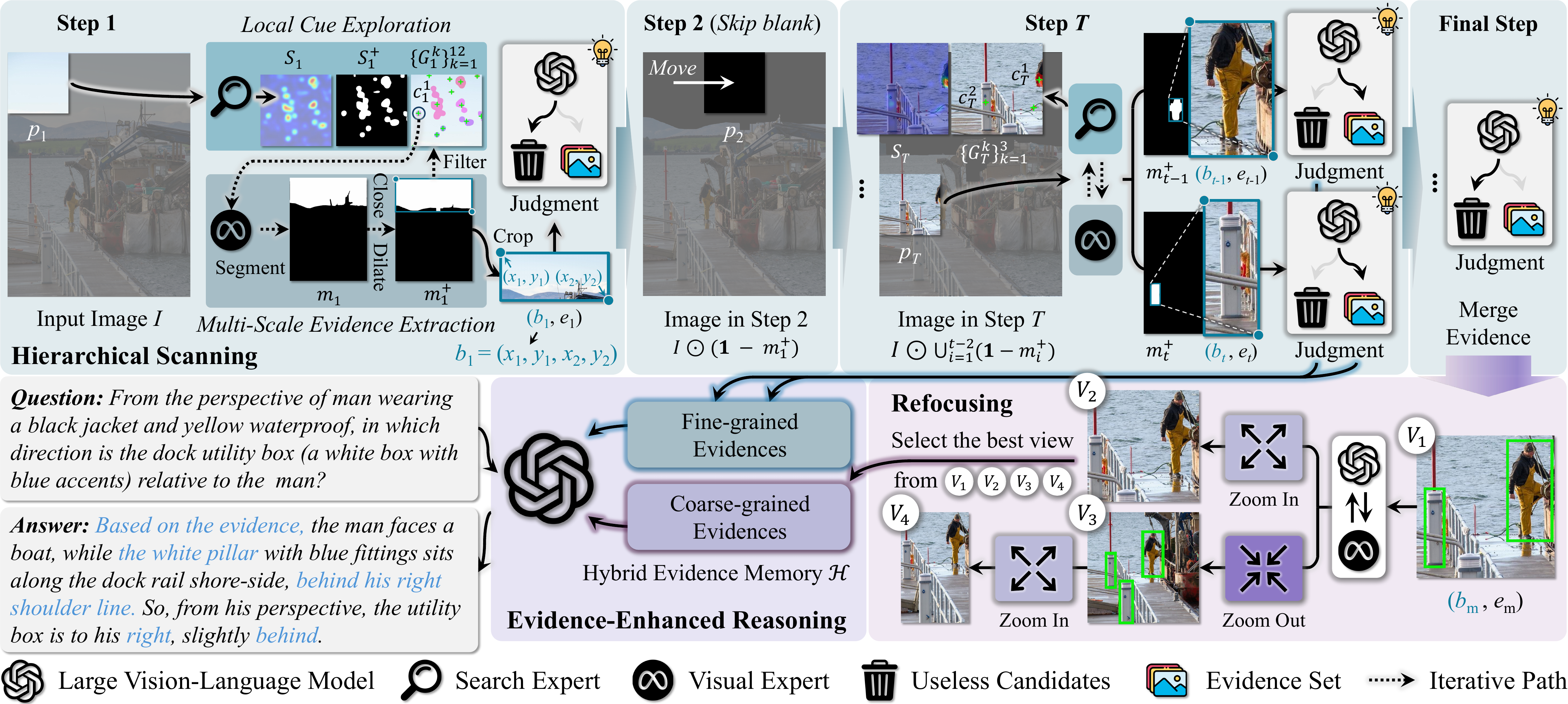}
	\caption{Overall architecture of \textbf{DeepScan}:
    Hierarchical Scanning progressively recovers visual evidence from \emph{in-patch} cues formulated as point-based proxies using Local Cue Exploration and Multi-Scale Evidence Extraction, \eg, $c^1_{T} \mapsto e_{t-1}$ in step $T$; Refocusing further refines the surrounding context for the fused evidence via interactions between LVLMs and visual experts; Evidence-Enhanced Reasoning leverages a Hybrid Evidence Memory to provide multi-granular information to the LVLM, enabling detailed yet comprehensive answers.
    }
	\label{fig:3}
    \vspace{-2mm}
\end{figure*}

\subsection{Visually Grounded Reasoning}
Recent agents with \emph{``think-with-images''} capabilities (\eg, GPT-o3~\cite{o3}) perform visually grounded reasoning that localizes and emphasizes the critical visual content during inference through dynamic image manipulation~\cite{xie2024large,su2025thinking,o3}. Building on this, subsequent works employ SFT and reinforcement learning (RL) with visual rewards~\cite{wang2025vgr,zheng2025deepeyes,wang2025traceable,zhang2025thyme,su2025openthinkimg,lai2025mini}, or curiosity-driven objectives~\cite{su2025pixelreasoner} to reproduce this ability in general LVLMs; SEAL~\cite{wu2024vstar} further integrates explicit evidence localization and visual memory. However, these approaches are costly and challenging to generalize to different architectures and model scales. To overcome this, a parallel line explores training-free paradigms that leverage external visual experts~\cite{li2025dyfo,an2025mitigating,You2024Ferret,Zhang2025PSALM}, tree search~\cite{shen2024zoomeye}, or self-refinement~\cite{yu2025zoom} to enhance LVLMs by ``grounding-then-answering.'' Despite these advances, most of them rely on a \emph{coarse-to-fine} grounding strategy, which is sensitive to noisy context and semantically similar objects, resulting in inaccurate localization. By contrast, we propose Hierarchical Scanning, a \emph{bottom-up} grounding paradigm that progressively recovers visual evidence from associated local cues via point-based proxies, enabling robust localization of fine-grained visual evidence in complex scenes.
\section{Methodology}
\label{Sec:3}

\subsection{Overview}
\label{subsec:overview}

To equip LVLMs with visually grounded reasoning, we propose \textbf{DeepScan} that localizes, recalibrates, and integrates visual evidence to produce more accurate and interpretable answers, as shown in~\cref{fig:3}. To enhance fine-grained perception, DeepScan augments LVLMs with two \emph{plug-and-play} models, \ie, a search expert and a visual expert. Formally, given an image $I\in\mathbb{R}^{H\times W\times3}$ with a question $q$, the search expert highlights the potential cues in a local patch $p\in I$ via its attention map $S$ induced by GradCAM~\cite{Selvaraju2017GradCAM}:
\begin{tightEq}
\begin{equation}
    S =\textsc{Search}(p,q)\in\mathbb{R}^{h\times w},\quad p\in\mathbb{R}^{h\times w\times3}.
\end{equation}
\end{tightEq}
By contrast, the visual expert exposes two primitives to indicate the image-level evidence: segmentation from a point prompt $c=(x,y)\in I$ and detection from the question $q$:
\begin{tightEq}
\begin{equation}
    m=\textsc{Segment}(I,c),\quad \mathcal{B}=\textsc{Detect}(I,q),
\end{equation}
\end{tightEq}
where $m$ denotes the evidence mask, and $\mathcal{B}$ represents the union of evidence bounding boxes. In this way, the LVLM can generate a well-grounded answer for the question using the image regions enclosed by $m$ or $\mathcal{B}$. This pipeline is general and training-free, improving LVLMs across architectures and model scales without adaptation cost.

\subsection{Hierarchical Scanning}
\label{sec:hs}
To suppress the effects of irrelevant visual content, a natural idea is to partition the image into patches and perform a patch-wise evidence extraction. However, this poses a fundamental challenge: when evidence spans multiple patches, conventional proxies (\eg, region proposals~\cite{su2025pixelreasoner,zheng2025deepeyes,wang2025traceable,zhang2025thyme}, detection boxes~\cite{li2025dyfo,qian2025zoomer}, or textual descriptions~\cite{wang2025hrbench}) struggle to represent the resulting incomplete evidence.

\begin{figure}[t]
	\centering
	\includegraphics[width=1\linewidth]{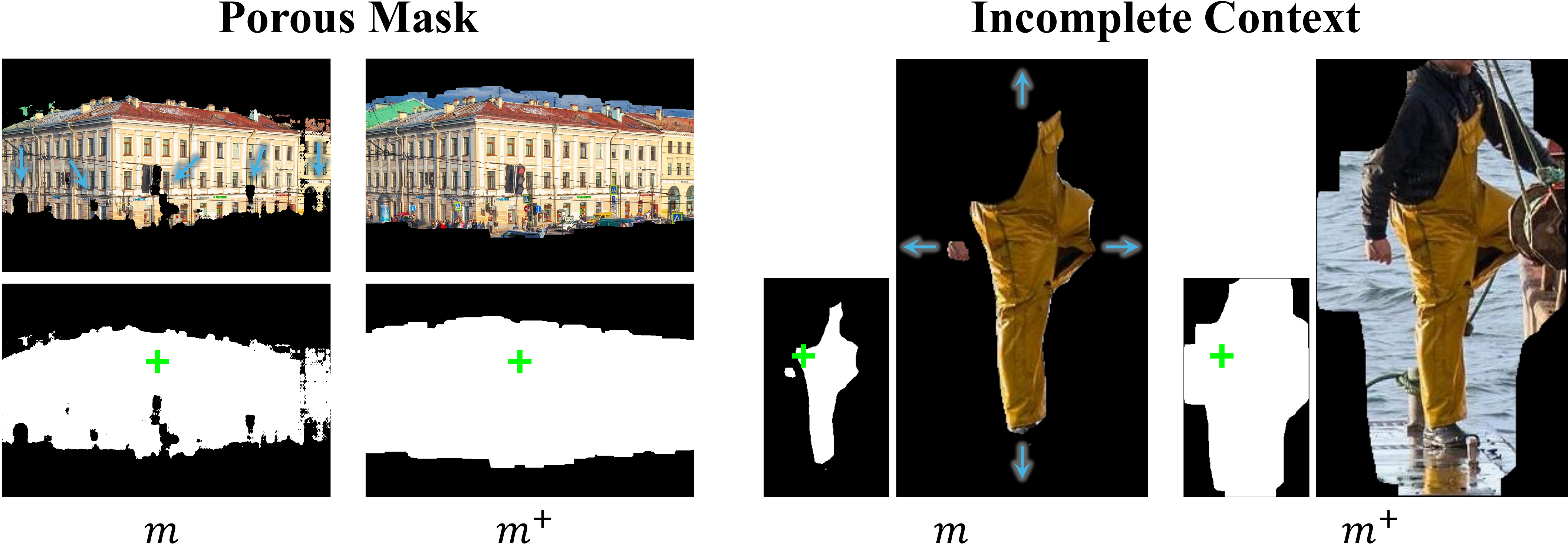}
	\caption{Illustration of %deviations in point-prompt segmentation and their correction via 
    morphological post-processing. \textcolor{green}{\textbf{+}} marks the point-based proxies; \(m\) and \(m^+\) denote evidence masks before and after post-processing, showing improved robustness.}
	\label{fig:4}
    \vspace{-2mm}
\end{figure}

To address this issue, Hierarchical Scanning introduces novel \emph{point-based proxies} to bridge two well-designed processes, \ie, \emph{patch-wise} Local Cue Exploration and \emph{image-level} Multi-scale Evidence Extraction, thereby enabling a bottom-up visual grounding paradigm. 

\emph{Local Cue Exploration.}
Given a patch \(p \in \mathbb{R}^{h\times w\times 3}\) from the input image \(I\) and a question \(q\), we first apply the search expert and Otsu’s method~\cite{otsu1975threshold} to produce potential cue regions indicated by a high-value attention mask \(S_p^{+}\):
\begin{tightEq}
\begin{equation}
    S_p^{+} = \mathbb{I}\big(S_p \ge T_p^\star\big)\in\{0,1\}^{h\times w},\ 
     T_p^\star = \textsc{Otsu}(S_p),
\end{equation}
\end{tightEq}
where \(S_p = \textsc{Search}(p, q) \in \mathbb{R}^{h\times w}\). We then formulate the \emph{cues} \(\{G_p^k\}_{k=1}^{K}\) as the connected components in \(S_p^{+}\). 
To associate these cues with evidence, we represent each cue by an interior point derived from its geometric and semantic information, enabling efficient evidence retrieval via point-prompt segmentation. For any interior location $c\in G$, the geometric term is its distance to the cue boundary $\partial G$:
\begin{tightEq}
\begin{equation}
d(c,\partial G)=\inf_{\gamma\in  \partial G} \|c-\gamma \|_2.
\label{equ:4}
\end{equation}
\end{tightEq}
Considering only~\cref{equ:4} can yield multiple candidates for complex cues (\eg, U-shaped regions); we therefore incorporate the attention scores to disambiguate and bias the evidence proxies \(\mathcal{C}_p\) toward semantically significant regions:
\begin{tightEq}
\begin{equation}
\mathcal{C}_p=\Big\{c_p^k \mid c_p^k{=}\argmax_{c\in G_p^k} \tilde{S}_p(c)\,\tilde{d}(c,\partial G_p^k),\ |G_p^k|{\ge}\tau\Big\},
\label{equ:12}
\end{equation}
\end{tightEq}
where $\tau$ is a threshold that filters out spurious small cues.
$\tilde{S}_p(c)$ is the normalized attention score at $c$, and $\tilde{d}(c,\partial G_p^k)$ is the normalized distance-to-boundary at $c$. 
Finally, we lift these \emph{in-patch} proxies \(\mathcal{C}_p\) to the image coordinates for \(\mathcal{C}_p'\).

\begin{figure}[t]
	\centering
	\includegraphics[width=0.98\linewidth]{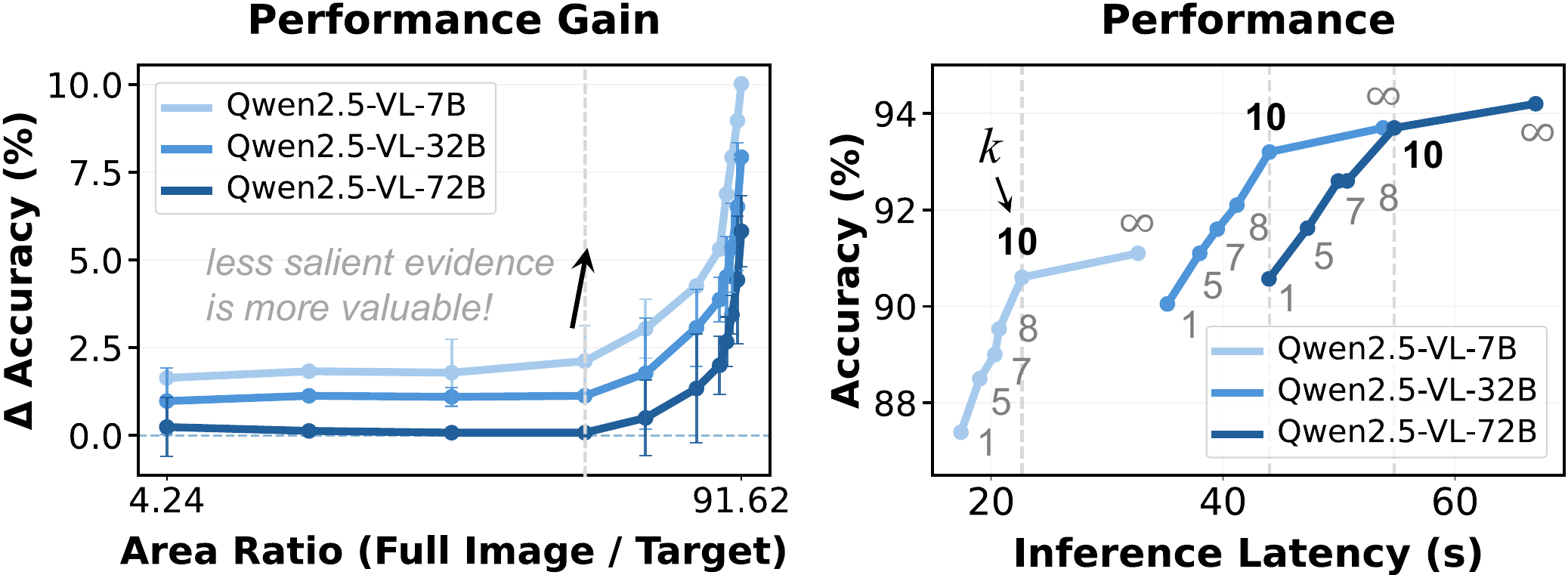}
    \vspace{-1.1mm}
	\caption{Analysis of (\emph{left}) performance gain \wrt target area ratio and (\emph{right}) performance-latency trade-off on V*, where $\infty$ mark denotes the case \emph{without} explicit truncation of candidate count $k$.}
	\label{fig:5}
    \vspace{-3.8mm}
\end{figure}

\emph{Multi-scale Evidence Extraction.} Given a proxy \(c^\prime_p \in \mathcal{C}^\prime_{p}\) obtained from image $I$, we employ the visual expert to recover the evidence mask $m$ by point-prompt segmentation:
\begin{tightEq}
\begin{equation}
    m = \textsc{Segment}(I, c^\prime_p)\!\in\! \{0,1\}^{H \times W},\ I\! \in\! \mathbb{R}^{H \times W \times 3}.
\end{equation}
\end{tightEq}
In particular, although the mask \(m\) captures image-level evidence, the single-point prompt has limited expressiveness; as a result, \(m\) often contains artifacts, \eg, interior holes or incomplete surrounding context, as shown in~\cref{fig:4}. To ensure the evidence integrity, we adopt morphological post-processing and produce the enhanced mask \(m^{+}\):
\begin{tightEq}
\begin{equation}
    m^+= \bigl(m \bullet \mathcal{K}\bigr) \,\oplus\, \mathcal{S}_r, \qquad m^{+}\in \{0,1\}^{H \times W},
\end{equation}
\end{tightEq}
where \(\bullet\) denotes \emph{closing} with a flat kernel \(\mathcal{K}\) to seal interior holes, and \(\oplus\) denotes \emph{dilation} with a disk kernel \(\mathcal{S}_r\) to extend the mask outward.
To avoid repeated extraction, we filter any proxies that fall inside the same evidence mask:
\begin{tightEq}
\begin{equation}
    \mathcal{C}^\prime_p \leftarrow \left\{\, c \in \mathcal{C}^\prime_p \;\middle|\; m^+(c) = 0 \right\},
\end{equation}
\end{tightEq}
where \(m^+(c)\) denotes the value of \(m^+\) at the location indicated by proxy \(c\).
The evidence candidate $e$ is then cropped from the minimal enclosing region $b$ of \(m^{+}\).
% \begin{tightEq}
% \begin{equation}
%     b = \textsc{BBox}(m^{+}), \qquad e = \textsc{Crop}(I, b).
% \end{equation}
% \end{tightEq}
Motivated by~\cite{li2025dyfo,yu2025zoom,wang2025hrbench}, we query the LVLM to make a binary judgment of the evidence $e$ (detailed in Supp.~\ref{supp:detials}) and, if affirmed, update the evidence set \(\mathcal{E} \leftarrow \mathcal{E} \cup \{(b,\,e)\}\).

Finally, we maintain a global visited mask over the examined evidence, \ie, $I\gets I\odot (1-m^+)$, and bypass the masked region to reduce redundant computation.

\noindent\textbf{Heuristic Acceleration.}
As shown in \cref{fig:5}(\emph{left}), localizing less salient evidence leads to a significantly higher performance gain for LVLMs. This is because large evidence regions are often detected by the LVLM without explicit grounding. Therefore, we pre-filter candidates by area and evaluate only the top-$k$ smallest regions, which limits the number of LVLM evaluations to $k$ and reduces the visual-token cost. Empirically, even $k=1$ preserves about 96\% of the maximum achievable performance while yielding roughly a $2\times$ speedup, as presented in~\cref{fig:5}(\emph{right}).

\begin{figure}[t]
	\centering
	\includegraphics[width=1\linewidth]{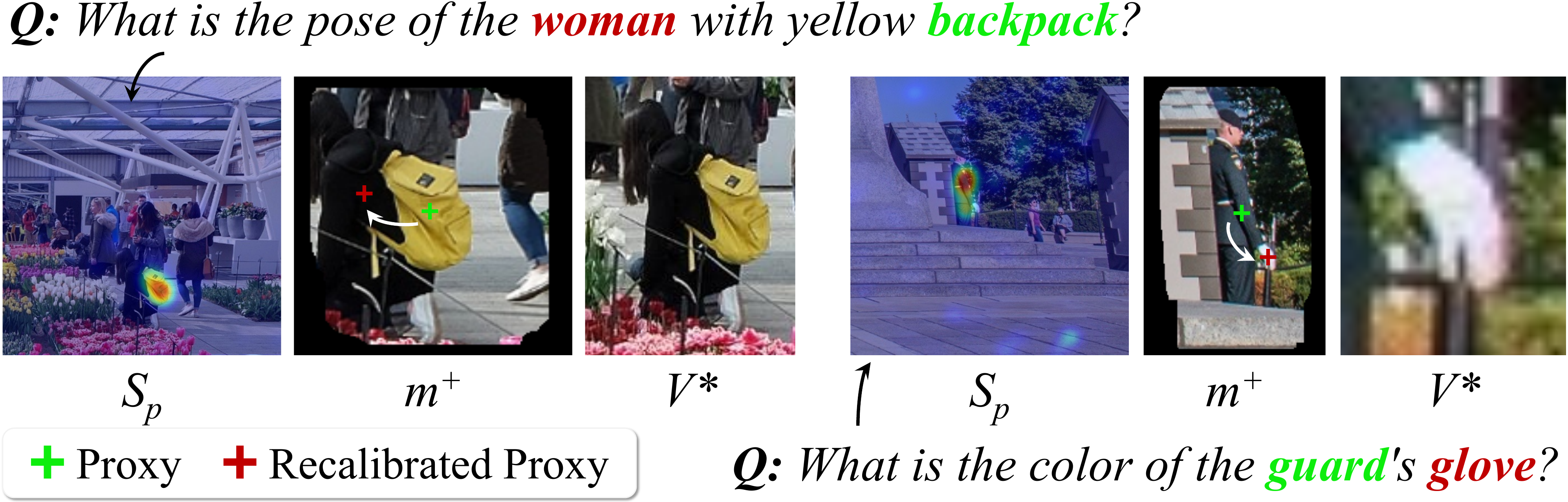}
	\caption{Illustration of Refocusing, where $V^*$ denotes its results. It reveals that Refocusing recalibrates the proxy misalignment by adaptively completing (\emph{left}) or further amplifying (\emph{right}) evidence.}
	\label{fig:6}
    \vspace{-2.mm}
\end{figure}

\subsection{Refocusing}
\label{sec:refocus}
While Hierarchical Scanning effectively recovers evidence via point-based proxies, precise proxy placement becomes challenging when multiple visual elements are spatially adjacent, often yielding insufficient or excessive surrounding context, as shown in~\cref{fig:6}. To address this, we introduce Refocusing, a collaborative search paradigm that integrates the LVLM with the visual expert to identify an evidence-centric view with an optimal surrounding context window.

\emph{Initialization.}
Given an evidence set $\mathcal{E}$ collected by Hierarchical Scanning in image $I$, we initialize the search with the aggregated evidence, \ie, $V_1=\textsc{Crop}(I, b_{\rm m})$, where $b_{\rm m}$ is the minimum bounding box enclosing all evidence. 

\emph{States \& Successor Function.}
Search states are defined as neighborhoods of $V_1$ obtained by varying the surrounding context. We design two actions to generate candidate states:
\begin{enumerate}[leftmargin=*,
                  itemsep=0pt,
                  parsep=0pt,
                  topsep=0pt,
                  partopsep=0pt]
    \item[$\diamondsuit$] \emph{Zoom-In.} Narrow the surrounding context by cropping $V$ to the union of detections conditioned on query $q$:
    \begin{tightEq}
    \begin{equation}
        \textsc{In}(V,q)= \textsc{Crop}(V,\, \textsc{Detect}(V,q)).
    \end{equation}
    \end{tightEq}

    \item[$\diamondsuit$] \emph{Zoom-Out.} Enlarge the surrounding context by isotropically expanding $V$ with scale $s>1$ \wrt its center:
    \begin{tightEq}
    \begin{equation}
        \textsc{Out}(V,s) = \textsc{Crop}(I,\, \textsc{ScaleBbox}(V,s)).
    \end{equation}
    \end{tightEq}
\end{enumerate}

\emph{Selection Policy.}
Motivated by \cref{fig:5}(\emph{left}), we prefer the smallest view that still contains all evidence required to answer the question. Accordingly, we define an LVLM-based reward and select the state with the highest score:
\begin{tightEq}
\begin{equation}
    R(V) \;=\; \mathbb{I}_{V\leadsto q}\cdot HW/hw,\quad V\in\mathbb{R}^{h\times w\times3}.
\label{equ:14}
\end{equation}
\end{tightEq}
Here, $\mathbb{I}_{V\leadsto q}$ is an indicator equal to $1$ if $V$ contains sufficient evidence for answering $q$, and $0$ otherwise. In practice, $\mathbb{I}_{V\leadsto q}$ is obtained from a one-shot binary judgment by the LVLM over the objects in $\mathcal{E}$ (details in Supp.~\ref{supp:detials}).

\emph{Search Space Design.}
Unlike existing methods~\cite{shen2024zoomeye,li2025dyfo} that search the entire image, Hierarchical Scanning provides a well-initialized view $V_1$ for Refocusing, which motivates the following design principles to prune the search space:
\begin{enumerate}[leftmargin=*, itemsep=0pt, parsep=0pt, topsep=0pt, partopsep=0pt]
    \item[$\diamondsuit$]
    The optimal evidence view lies in a neighborhood of $V_1$; thus, \emph{Zoom-In} at $V_1$ is empirically idempotent:
    \begin{tightEq}
    \begin{equation}
        \textsc{In}(\textsc{In}(V_1,q),q) \;=\; \textsc{In}(V_1,q)\;=\;evidence.
    \end{equation}
    \end{tightEq}

    \item[$\diamondsuit$]
    $V_1$ covers a small fraction of $I$; thus, \emph{Zoom-Out} expands $V_1$ without boundary clipping. Hence, for any $s_1,s_2 \!\ge\! 1$,
    \begin{tightEq}
    \begin{equation}
        \textsc{Out}(\textsc{Out}(V_1,s_1),s_2) \;=\; \textsc{Out}(V_1,s_1s_2).
    \end{equation}
    \end{tightEq}

    \item[$\diamondsuit$]
    If $V_1$ already contains complete evidence, \ie, $\mathbb{I}_{V_1 \leadsto q}{=}1$, then
    $\mathbb{I}_{\textsc{In}(V_1,q)\leadsto q}=\mathbb{I}_{\textsc{Out}(\textsc{In}(V_1,q),s)\leadsto q}=1$, and thus
    \begin{tightEq}
    \begin{equation}
        R(\textsc{Out}(\textsc{In}(V_1,q),s))\le R(\textsc{In}(V_1,q)).
    \end{equation}
    \end{tightEq}
    Otherwise (\ie, $\mathbb{I}_{\textsc{In}(V_1,q)\leadsto q}=\mathbb{I}_{V_1\leadsto q}=0$):
    if \emph{Zoom-Out} fails to restore the missing context for $\textsc{In}(V_1,q)$, then
    \begin{tightEq}
    \begin{equation}
    R(\textsc{Out}(\textsc{In}(V_1,q),s))=0;
    \end{equation}
    \end{tightEq}
    if \emph{Zoom-Out} can restore the context for $\textsc{In}(V_1,q)$, then it can also restore the context directly from $V_1$, and thus
    \begin{tightEq}
    \begin{equation}
        R(\textsc{Out}(\textsc{In}(V_1,q),s))\le R(\textsc{In}(\textsc{Out}(V_1,s),q)).
    \end{equation}
    \end{tightEq}
    Thus, the state $\textsc{Out}(\textsc{In}(V_1,q),s)$ can be omitted from the search space without affecting global optimality.
\end{enumerate}
Based on these principles, we construct the search space via only a depth-2 expansion of $V_1$ and consider a \emph{concise yet behaviorally complete} view set $\mathcal{V}=\{V_1, V_2, V_3, V_4\}$:
\begin{tightEq}
\begin{equation}
V_2=\textsc{In}(V_1,q),\
V_3=\textsc{Out}(V_1,s),\
V_4=\textsc{In}(V_3,q),
\end{equation}
\end{tightEq}
where the scale $s$ is set to $1.5$ via a grid search. We traverse $\mathcal{V}$ in depth-first order and greedily select the best view.

\subsection{Evidence-Enhanced Reasoning}
\label{sec:eer}
During the preceding stages, we build a hybrid evidence memory \(\mathcal{H}\) that stores fine-grained evidence from Hierarchical Scanning and coarse-grained views from Refocusing:
\begin{tightEq}
\begin{equation}
    \mathcal{H}
    = \bigl\{\, e,\, V^* \ \big|\ (b,e)\in\mathcal{E},\ 
       V^* {=} \operatorname*{\arg\max}_{V\in\mathcal{V}} R(V) \bigr\}.
\end{equation}
\end{tightEq}
This hybrid memory $\mathcal{H}$ is then materialized as an ordered \emph{multi-image prompt} $[e_1,\dots,V^*]$.
Leveraging this, LVLMs can resolve object attributes from fine-grained evidence and infer relations from coarse-grained views, yielding more comprehensive and accurate answers \(\mathcal{A} = \textsc{Reason}(\mathcal{H},\, q)\).

\begin{table}[t]
    \centering\footnotesize
    % \centering\small    
    \setlength{\tabcolsep}{2.3pt}
    \renewcommand{\arraystretch}{0.6}
    \caption{
    Comparisons on V* Bench and HR-Bench, where all the baselines are developed from Qwen2.5-VL-7B, and the RL-based methods are marked by \textcolor{gray}{gray}. Results$^\dagger$ are self-collected. 
    } 
    \vspace{-2.mm}
    \begin{tabular}{l | ccc ccc ccc}
    \toprule
    & \multicolumn{3}{c}{V*~\cite{wu2024vstar}} & \multicolumn{3}{c}{HR-4K~\cite{wang2025hrbench}} & \multicolumn{3}{c}{HR-8K~\cite{wang2025hrbench}} \\
    \cmidrule(lr){2-4}
    \cmidrule(lr){5-7}
    \cmidrule(lr){8-10}
    & Avg & \emph{Att} & \emph{Spa} & Avg & \emph{Sin} & \emph{Cro} & Avg & \emph{Sin} & \emph{Cro} \\ 
    \midrule
    \multicolumn{10}{c}{\textbf{General Large Vision-Language Models}} \\
    \midrule
    GPT-4o-1120~\cite{gpt4o} & 66.0 & -- & -- & 59.0 & 70.0 & 48.0 & 55.5 & 62.0 & 49.0 \\
    \midrule
    LLaVA-OV-7B\cite{li2024llavaov} & 70.7 & 73.0 & 60.5 & 64.3 & 74.8 & 53.8 & 59.8 & 65.3 & 54.3 \\
    LLaVA-OV-72B & 73.8 & 80.9 & 63.2 & 66.3 & 76.5 & 56.0 & 60.9 & 68.8 & 53.0 \\
    InternVL3-8B\cite{zhu2025internvl3} & 72.3 & 73.0 & 71.1 & 70.8 & 79.3 & 62.3 & 62.0 & 64.3 & 59.8 \\
    InternVL3-38B & 77.5 & 77.4 & 77.6 & 76.3 & 83.5 & 69.0 & 67.0 & 71.3 & 62.8 \\
    InternVL3-78B & 76.4 & 75.7 & 77.6 & 75.5 & 84.5 & 66.5 & 67.3 & 71.8 & 62.8 \\
    Qwen2.5VL-7B\cite{bai2025qwen25vl} & 74.3 & 77.4 & 69.7 & 72.1 & 88.8 & 55.5 & 68.8 & 83.5 & 54.0 \\
    Qwen2.5VL-32B & 85.9 & 83.5 & 89.5 & 74.8 & 89.3 & 60.3 & 71.6 & 86.5 & 56.8 \\
    Qwen2.5VL-72B & 84.8 & 90.8 & 80.9 & 79.4 & 88.8 & 70.0 & 76.3 & 84.3 & 68.3 \\
    \midrule
    \multicolumn{10}{c}{\textbf{Visually Grounded Reasoning Models}} \\
    \midrule
    \textcolor{gray}{PixelReasoner}\cite{su2025pixelreasoner} & 80.6 & 83.5 & 76.3 & 72.9 & 86.0 & 60.3 & 66.9 & 80.0 & 54.3 \\
    \textcolor{gray}{DeepEyes}\cite{zheng2025deepeyes} & \cellcolor{blue!5}{90.0} & \cellcolor{blue!5}{92.1} & \cellcolor{blue!15}{86.8} & \cellcolor{blue!5}{75.1} & \cellcolor{blue!15}{91.3} & 59.0 & \cellcolor{blue!15}{72.6} & \cellcolor{blue!5}{86.8} & \cellcolor{blue!15}{58.5} \\
    \textcolor{gray}{Thyme-VL}\cite{zhang2025thyme}        & 82.2 & 83.5 & 80.3 & \cellcolor{blue!15}{77.0} & \cellcolor{blue!5}{91.0} & \cellcolor{blue!15}{63.0} & 72.0 & 86.5 & \cellcolor{blue!5}{57.5} \\
    \textcolor{gray}{TreeVGR}$^\dagger$\cite{wang2025traceable} & 85.9 & 86.1 & \cellcolor{blue!5}{85.5} & 72.7 & 89.5 & \cellcolor{blue!5}{61.5} & 69.8 & 84.4 & 57.2 \\
    \midrule
    ZoomRefine$^\dagger$\cite{yu2025zoom} & 82.2 & 85.3 & 77.6 & 71.5 & 88.5 & 55.3 & 68.6 & 83.9 & 54.0 \\  
    Dyfo$^\dagger$\cite{li2025dyfo} & 84.3 & 82.6 & \cellcolor{blue!15}{86.8} & 71.3 & 89.2 & 53.5 & 69.8 & 86.5 & 53.2 \\   
    \textbf{DeepScan} & \cellcolor{blue!15}{90.6} & \cellcolor{blue!15}{93.0} & \cellcolor{blue!15}{86.8} & \cellcolor{blue!5}{75.0} & 90.1 & 59.7 & \cellcolor{blue!5}{72.4}  & \cellcolor{blue!15}{87.2} & \cellcolor{blue!5}{57.6} \\
    \vs {\scriptsize Qwen2.5-VL-7B} & \up{16} & \up{16} & \up{17} & \up{2.8} & \up{1.3} & \up{4.2} & \up{3.6} & \up{3.7} & \up{3.6} \\
    \bottomrule
    \end{tabular}
    \vspace{-3mm}
    \label{tab:1}
\end{table}

\begin{table*}[t]
    \centering\footnotesize
    \setlength{\tabcolsep}{2pt}
    \renewcommand{\arraystretch}{0.55}
    \caption{Comparison with state-of-the-art alternatives on TreeBench~\cite{wang2025traceable}. All the baselines are developed from Qwen2.5-VL-7B, and the RL-based methods are marked by \textcolor{gray}{gray}. Results$^\dagger$ are self-collected. \textbf{DeepScan} \textit{achieves competitive results compared to SOTAs.}}
    \vspace{-2mm}
    \begin{tabular}{l | ccccccccccccc}
    \toprule
    & & & \rotatebox{25}{Attributes} & \rotatebox{25}{Material} & \rotatebox{25}{Phy. State} & \rotatebox{25}{Obj. Retr.} & \rotatebox{25}{OCR} & \rotatebox{25}{Per. Trans.} & \rotatebox{25}{Ordering} & \rotatebox{25}{Con. \& Oc.} & \rotatebox{25}{Spa. Cont.} & \rotatebox{25}{Comparison} \\
    \cmidrule(lr){4-8}
    \cmidrule(lr){9-13}
    & Overall & mIoU & \multicolumn{5}{c}{Perception} & \multicolumn{5}{c}{Reasoning} \\
    \midrule
    \multicolumn{13}{c}{\textbf{General Large Vision-Language Models}} \\
    \midrule
    Gemini-2.5-Flash-0520~\cite{gemini-2.5-flash} & 45.9 & -- & 48.3 & 53.9 & 69.6 & 68.8 & 75.0 & 15.3 & 19.3 & 56.1 & 72.4 & 43.2 \\
    GPT-4o-1120~\cite{gpt4o} & 46.9 & -- & 51.7 & 61.5 & 65.2 & 43.8 & 69.1 & 18.8 & 38.6 & 48.8 & 72.4 & 43.2 \\
    Gemini-2.5-Pro-0605~\cite{gemini-2.5-pro} & 54.1 & -- & 51.7 & 61.5 & 56.5 & 75.0 & 83.8 & 20.0 & 36.8 & 65.9 & 86.2 & 54.6 \\
    GPT-o3-0416~\cite{o3} & 54.8 & -- & 69.0 & 69.2 & 65.2 & 68.8 & 79.4 & 22.4 & 38.6 & 61.0 & 86.2 & 50.0 \\ 
    \midrule
    LLaVA-OneVision-7B~\cite{li2024llavaov} & 37.3 & -- & 55.2 & 53.8 & 56.5 & 50.0 & 32.4 & 21.2 & 22.8 & 41.5 & 72.4 & 36.4 \\
    LLaVA-OneVision-72B & 40.5 & -- & 62.1 & 53.8 & 65.2 & 62.3 & 36.8 & 12.9 & 28.1 & 53.7 & 65.5 & 47.7 \\
    Qwen2.5-VL-7B~\cite{bai2025qwen25vl} & 37.0 & -- & 55.2 & 53.8 & 56.5 & 62.5 & 27.9 & 20.0 & 35.1 & 39.0 & 44.8 & 43.2 \\
    Qwen2.5-VL-32B & 42.5 & -- & 51.7 & 53.8 & 69.6 & 62.5 & 54.4 & 16.5 & 33.3 & 46.3 & 62.1 & 38.6 \\
    Qwen2.5-VL-72B & 42.2 & -- & 65.5 & 69.2 & 56.5 & 56.3 & 48.5 & 11.8 & 33.3 & 51.2 & 72.4 & 38.6 \\
    InternVL3-8B~\cite{zhu2025internvl3} & 38.8 & -- & 51.7 & 69.2 & 56.5 & 56.3 & 33.7 & 21.2 & 24.6 & 39.0 & 72.4 & 43.2 \\
    InternVL3-38B & 42.0 & -- & 51.7 & 61.5 & 52.2 & 68.8 & 51.5 & 12.9 & 33.3 & 56.1 & 65.5 & 38.6 \\
    InternVL3-78B & 46.4 & -- & 62.1 & 61.5 & 52.2 & 68.8 & 52.9 & 16.5 & 33.3 & 61.0 & 86.2 & 45.5 \\
    \midrule
    \multicolumn{13}{c}{\textbf{Visually Grounded Reasoning Models}} \\
    \midrule
    \textcolor{gray}{DeepEyes}~\cite{zheng2025deepeyes} & 37.5 & 30.0 & \cellcolor{blue!15}{62.1} & 53.8 & \cellcolor{blue!15}{65.2} & \cellcolor{blue!15}{68.8} & \cellcolor{blue!5}{51.5} & 11.8 & 24.6 & 36.6 & \cellcolor{blue!15}{51.7} & \cellcolor{blue!15}{47.7} \\
    \textcolor{gray}{Pixel-Reasoner}~\cite{su2025pixelreasoner} & 39.0 & \cellcolor{blue!5}{35.7} & \cellcolor{blue!5}{58.6} & \cellcolor{blue!5}{61.5} & \cellcolor{blue!15}{65.2} & 50.0 & 48.5 & 14.1 & 31.6 & 39.0 & 44.8 & 40.9 \\
    \textcolor{gray}{TreeVGR}$^\dagger$ \cite{wang2025traceable} & \cellcolor{blue!5}{41.0} & 31.8 & 55.2 & \cellcolor{blue!5}{61.5} & \cellcolor{blue!15}{65.2} & 50.0 & \cellcolor{blue!15}{61.7} & 15.3 & 24.6 & \cellcolor{blue!15}{46.3} & 44.8 & 40.9 \\
    \midrule
    ZoomRefine$^\dagger$~\cite{yu2025zoom} & 38.0 & -- & 48.3 & \cellcolor{blue!5}{61.5} & 56.5 & \cellcolor{blue!5}{62.5} & 39.7 & \cellcolor{blue!5}{18.8} & 29.8 & \cellcolor{blue!15}{46.3} & 44.8 & 38.6 \\
    Dyfo$^\dagger$~\cite{li2025dyfo} & 39.3 & -- & \cellcolor{blue!5}{58.6} & \cellcolor{blue!15}{69.2} & 56.5 & \cellcolor{blue!5}{62.5} & 35.3 & \cellcolor{blue!15}{21.2} & \cellcolor{blue!5}{35.1} & 41.5 & 44.8 & 40.9\\
    \textbf{DeepScan} & \cellcolor{blue!15}{42.5} & \cellcolor{blue!15}{37.3} & \cellcolor{blue!15}{62.1} & \cellcolor{blue!15}{69.2} & \cellcolor{blue!5}{60.9} & \cellcolor{blue!15}{68.8} & 44.1 & \cellcolor{blue!15}{21.2} & \cellcolor{blue!15}{36.8} & \cellcolor{blue!5}{43.9} & \cellcolor{blue!5}{48.3} & \cellcolor{blue!5}{43.2} \\
    
    $\Delta$ \vs Qwen2.5-VL-7B & \up{5.5} & -- & \up{6.9} & \up{15.4} & \up{4.4} & \up{6.3} & \up{16.2} & \up{1.2} & \up{1.7} & \up{4.9} & \up{3.5} & \textcolor{gray}{$\pm$0.0} \\
    \bottomrule
    \end{tabular} %}
    \label{tab:2}
    \vspace{-2mm}
\end{table*}

% \newpage
\section{Experiments}
\subsection{Experimental Settings}
\noindent\textbf{Datasets \& Evaluation Metrics.} 
\textbf{V* Bench} contains 191 images with an average resolution of \(2246\times1582\) and emphasizes very small targets (mean area $< 0.05\%$). It covers two tasks: \emph{Direct Attribute} recognition (\emph{Att}, 115 samples) and \emph{Spatial Relationship} reasoning (\emph{Spa}, 76 samples). We report multiple-choice accuracy. \textbf{HR-Bench} is available at two resolutions (8K/4K). Each version contains 200 images, split evenly into \emph{Single-Instance Perception} (\emph{Sin}, 100 samples) and \emph{Cross-Instance Perception} (\emph{Cro}, 100 samples). We evaluate using multiple-choice accuracy with a cyclic-permutation protocol. \textbf{TreeBench}~\cite{wang2025traceable} is designed to assess ``\emph{thinking-with-images}'' capabilities via three principles: localization of traceable evidence, perception of subtle targets, and second-order reasoning. It provides 405 images with an average resolution of \(2152\times1615\) and reports both multiple-choice accuracy and localization quality (mIoU). 

\noindent\textbf{Baseline Methods.} We compare DeepScan with RL-based methods~\cite{su2025pixelreasoner,zheng2025deepeyes,wang2025traceable,zhang2025thyme}, training-free methods~\cite{yu2025zoom,li2025dyfo}, private models~\cite{o3,gpt4o,gemini-2.5-pro,gemini-2.5-flash}, and open-source general models~\cite{li2024llavaov,bai2025qwen25vl,zhu2025internvl3}. Further details are provided in Supp.~\ref{supp:result}.

\noindent\textbf{Implementation Details.} We adopt BLIP-ITM base~\cite{li2022blip,li-etal-2023-lavis} as the search expert and LangSAM~\cite{medeiros2024language} as the visual expert. To balance performance and latency, we set $k=10$ in practice. We evaluate DeepScan with 5 LVLMs: LLaVA-1.5-7B, Qwen2-VL-7B, and Qwen2.5-VL-7B / 32B / 72B. We query the LVLM to classify the question type (detailed in Supp.~\ref{supp:detials}) and dynamically set the patch size to \(576\times576\) for single-object and \(768\times768\) for multi-object scenarios. All the hyperparameters are listed in Supp.~\ref{supp:detials}.  All experiments were performed on 4$\,\times\,$NVIDIA L20 GPUs.

\subsection{Main Results} 
\noindent\textbf{Fine-grained Visual Understanding.}
As shown in \cref{tab:1}, DeepScan delivers substantial gains over vanilla Qwen2.5-VL-7B across all benchmarks, \eg, 16.3\% overall improvement on V*. Compared with the popular general LVLMs, DeepScan attains state-of-the-art performance. Noticeably, it even surpasses several 70B general models on perception tasks, such as the \emph{Attribute} subset in V* Bench and the \emph{Single} subset of HR-Bench, where evidence localization is crucial. Moreover, DeepScan consistently outperforms \emph{all the} training-free baselines, achieving 6.3\%, 3.6\%, and 2.6\% overall gains over DyFo on V* Bench and HR-Bench-4K/8K. Benefiting from the well-designed grounding pipeline, DeepScan remains competitive with leading RL-based methods, \eg, DeepEyes, \emph{without additional fine-tuning} of the LVLM, especially on perception tasks.

\noindent\textbf{General Visual Tasks.}  
In general, RL-based and training-free methods are comparable, as reported in \cref{tab:2}. Specifically, RL-based approaches have clear advantages on complex perception tasks (\eg, Physical State, OCR, \emph{etc}.), whereas on second-order reasoning tasks (\eg, Perspective Transform, Ordering, \emph{etc}.) they deliver trivial gains. We hypothesize that RL does not fundamentally strengthen LVLMs in visual reasoning; rather, it biases LVLMs toward perception behaviors. Furthermore, DeepScan shows strong perception ability, exceeding the RL-based DeepEyes, Pixel-Reasoner, and TreeVGR by 7.3, 1.6, and 5.5 mIoU, respectively, revealing the superiority of Hierarchical Scanning. Based on this, DeepScan attains the best overall results among not only training-free but also RL-based baselines, outperforming DeepEyes, Pixel-Reasoner, and TreeVGR by 5.0\%, 3.5\%, and 1.5\% without fine-tuning.

\begin{figure}[t]
	\centering
	\includegraphics[width=0.9\linewidth]{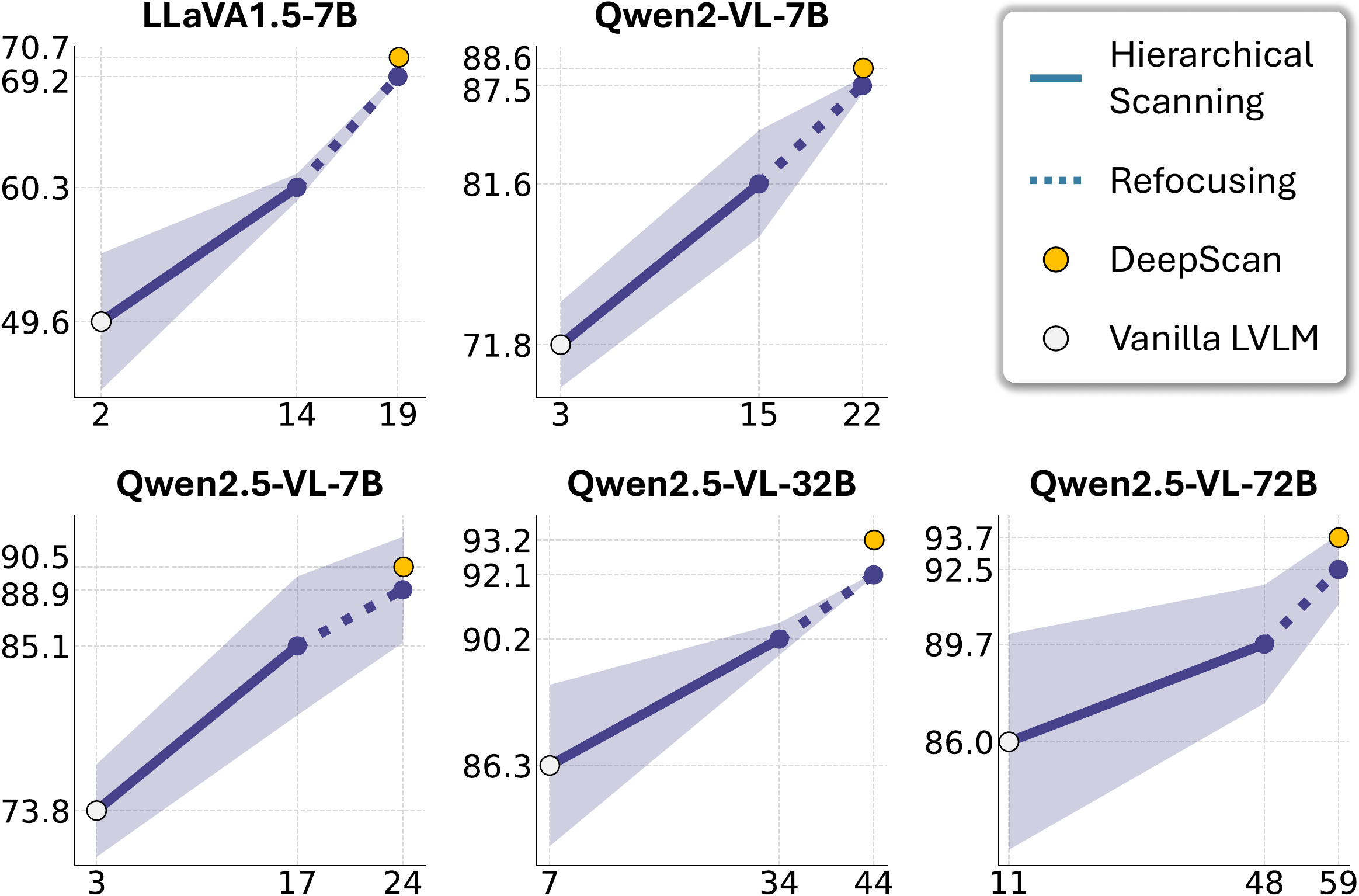}
	\caption{Ablations of DeepScan across LVLMs on V*, where the $y$-axis shows average performance (\%), the $x$-axis reports latency (s), and the shadow shows performance deviation across subsets.
    }
	\label{fig:8}
    \vspace{-2mm}
\end{figure}

\subsection{Ablation Study}
\label{sec:4.3}
\noindent\textbf{Framework.} 
We conduct ablations of the DeepScan framework in two aspects: its components and overall pipelines.
\noindent\emph{$\diamondsuit$ Components.} As presented in~\cref{fig:8}, DeepScan consistently improves diverse LVLMs across architectures and parameter scales without adaptation cost, demonstrating its strong generalization. Besides, \cref{tab:3} shows that the experts of different sizes deliver similar performance and exhibit comparable memory footprint and latency, indicating that DeepScan is insensitive to the expert scale.

\noindent\emph{$\diamondsuit$ Pipelines.} As illustrated in~\cref{fig:8}, all stages contribute to the improvement. Specifically, Hierarchical Scanning is the primary driver; Refocusing adds further gains while maintaining a favorable performance–latency trade-off. Building on them, Evidence-Enhanced Reasoning delivers additional improvements with negligible overhead. This demonstrates the completeness of the DeepScan framework.

\begin{table}[t]
    \centering\footnotesize
    \setlength{\tabcolsep}{4.pt}
    \renewcommand{\arraystretch}{0.7}
    \caption{Ablation of external experts on V*; Mem denotes the memory footprint of each DeepScan variant in gigabytes (G). Time reports the average per-sample inference latency in seconds (s).}
    \vspace{-2mm}
    \begin{tabular}{@{}p{15mm}@{}l|ccccc@{}}
    \toprule
    & & Overall & Attribute & Spatial & Mem ($\downarrow$) & Time ($\downarrow$)\\
    \midrule
    \multirow{2}{*}{\shortstack{BLIP-ITM}} & base           & \textbf{90.6} & \textbf{93.0} & {86.8} & 29.4G  & 24.5s \\
                                             & large          & {90.1}        & {92.2}        & \textbf{86.8} & 32.8G  & 25.8s \\
    \midrule    
    \multirow{3.5}{*}{LangSAM}               & small          & {89.5}        & \textbf{93.0} & 84.2  & 31.5G & 23.2s \\
                                             & base$^+$       & {89.5}        & 91.3          & \textbf{86.8} & 31.9G & 24.0s \\
                                             & large          & \textbf{90.6} & \textbf{93.0} & \textbf{86.8} & 32.8G & 24.5s \\
    \bottomrule
    \end{tabular}
    \vspace{-0.5mm}
    \label{tab:3}
\end{table}
\begin{table}[t]
    \centering\footnotesize
    \setlength{\tabcolsep}{4.5pt}
    \renewcommand{\arraystretch}{0.7}
    \caption{Ablation of Hierarchical Scanning on V*.}
    \vspace{-2mm}
    \begin{tabular}{l | cccc}
    \toprule
    & Overall & Attribute & Spatial & Time ($\downarrow$)\\
    \midrule
    Detection~\cite{li2025dyfo}   & 82.2 & 81.7 & 82.9 & 13.0s\\
    Hierarchical Scanning  & \textbf{90.6} & \textbf{93.0} & \textbf{86.8} & 24.5s\\
    % $\Delta$                & \up{8.4}    & \up{11.3} &\up{3.9} & \textcolor{red}{$\times1.87$}\\
    \midrule
    $w/o$ Post-Processing   & 87.4 & 89.6 & 85.5 & 32.1s\\
    % $\Delta$                & \down{3.2}    & \down{3.4} &\down{1.3} & \textcolor{red}{$\times1.31$} \\
    \bottomrule
    \end{tabular}
    \label{tab:4}
    \vspace{-0.5mm}
\end{table}
\begin{table}[t]
\centering
\footnotesize
\setlength{\tabcolsep}{3pt}
\renewcommand{\arraystretch}{0.7}
\caption{Ablation of the Local Cue Exploration on V*. \emph{left}: proxy type, where S and T denote semantic and topological information. \emph{right}: patch size, where S / M denotes single-/multi-object scenes.}
\vspace{-2mm}
\begin{subtable}[t]{0.56\linewidth}
\centering
\begin{tabular}{@{}l|cc|ccc@{}}
\toprule
Proxy      & S & T & Att & Spa \\
\midrule
Centroid         &  &\checkmark   & 84.3 & 80.3 \\
Chebyshev Center &  & \checkmark  & \textbf{91.3} & 82.9 \\
Attention Peak   &\checkmark  &   & 87.8 & \textbf{85.5} \\
\midrule
\textbf{Ours}       & \checkmark & \checkmark  & \textbf{93.0} & \textbf{86.8} \\

\bottomrule
\end{tabular}
\label{tab:5a}
\end{subtable}
\hfill
\begin{subtable}[t]{0.4\linewidth}
\centering
\begin{tabular}{@{}c|ccc@{}}
\toprule
\emph{S} / \emph{M} & Avg & Att & Spa \\
\midrule
384        & 87.4 & 90.4 & 82.9 \\
576        & 90.1 & \textbf{94.0} & 84.2 \\
768        & 88.5 & 88.7 & \textbf{88.2} \\
\midrule
576 / 768  & \textbf{90.6} & 93.0 & 86.8 \\
\bottomrule
\end{tabular}
\label{tab:5b}
\end{subtable}
\label{tab:5}
\vspace{-2mm}
\end{table}

\noindent\textbf{Hierarchical Scanning} As shown in~\cref{tab:4}, compared with the detection-based grounding method applied in Dyfo~\cite{li2025dyfo}, Hierarchical Scanning delivers obvious overall gains, with particularly large improvements on the \emph{Attribute} subset. 

\noindent\emph{$\diamondsuit$ Morphological Post-Processing.} As shown in~\cref{tab:4}, it fills interior holes in evidence masks; thus, it not only yields clear performance gains but markedly accelerates inference by preventing repeated processing of the same evidence. 

\noindent\emph{$\diamondsuit$ Evidence Proxy.} As shown in~\cref{tab:5}(\emph{left}), centroid-based proxies can fall outside U-shaped cues, which compromises evidence recovery and degrades the overall performance. In contrast, attention peak and Chebyshev center provide complementary performance gains across scenes, while our design~\cref{equ:12} fuses them and yields noticeably better results.

\noindent\emph{$\diamondsuit$ Patch Size.} As reported in~\cref{tab:5}(\emph{right}), smaller patches are more effective at suppressing context and yield better performance in single-object scenes, while larger patches are preferred in multi-object scenes because context provides valuable spatial relations. Accordingly, we condition patch size on the scene type to enhance overall performance.

\noindent\textbf{Refocusing.} We conduct ablations of refocusing in three aspects: action set, search-space design, and zoom-out scale.

\noindent\emph{$\diamondsuit$ Action Set.} \cref{tab:6}(\emph{left}) shows that both actions are indispensable: removing either action degrades performance, whereas using them jointly yields consistent gains.

\noindent\emph{$\diamondsuit$ Search Space Design.} 
We evaluate both search efficiency and search space completeness using \emph{search length}, defined as the mean number of expansions required to first reach the oracle-optimal state. 
In ablation, MCTS and A* enumerate all depth-2 reachable states and score each using \cref{equ:14}, whereas \emph{Refocusing} visits only the states in \(\mathcal{V}\).
As shown in \cref{tab:6}(\emph{right}), under the same expansion budget, \emph{Refocusing} exhibits a shorter search length, empirically supporting our design principles and indicating both the completeness of the search space and the efficiency of the selection policy.

\begin{figure}[t]
	\centering
	\includegraphics[width=1\linewidth]{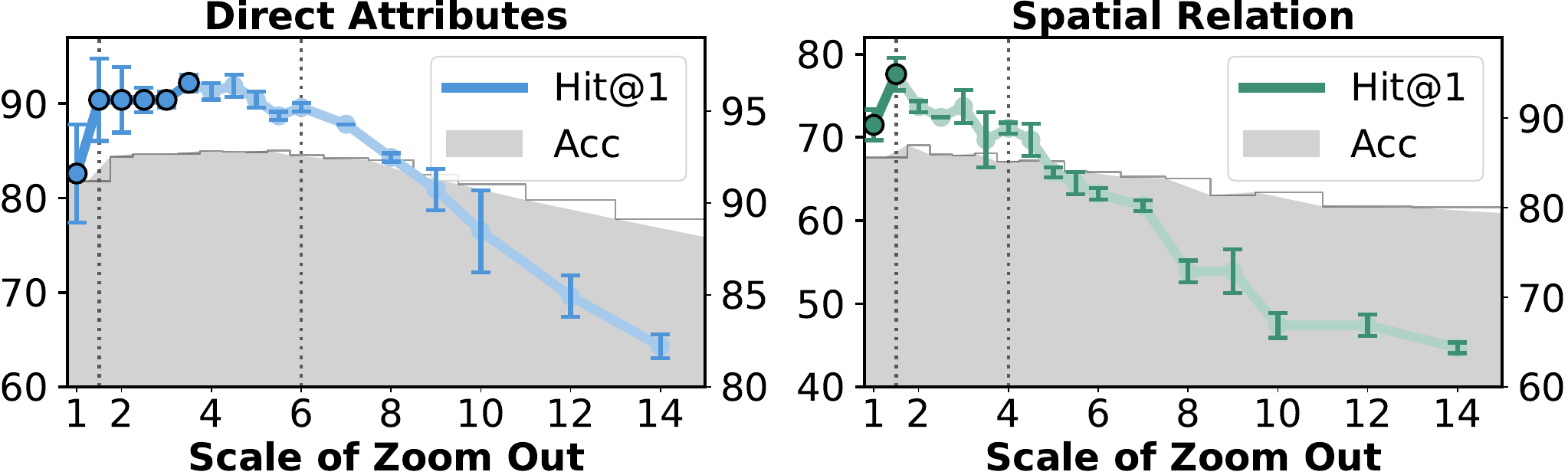}
	\caption{Ablation of the \emph{Zoom-Out} scale on V*, where the left y-axis shows Hit@1 of evidence detection on the zoomed-out view, while the right y-axis shows the accuracy (Acc) of DeepScan.
    } 
	\label{fig:9}
\end{figure}

\noindent\emph{$\diamondsuit$ Scale of Zoom-Out.} \cref{fig:9} shows that a moderate \emph{Zoom-Out} recalibrates evidence, \ie, higher Hit@1, and increases DeepScan accuracy, while overly large \emph{Zoom-Outs} bring excessive context and impair both grounding and reasoning. DeepScan shows greater tolerance to \emph{Zoom-Out} in \emph{Direct Attributes}, which is attributed to their smaller initial views.

\begin{table}[t]
\centering
\footnotesize
\setlength{\tabcolsep}{4pt}
\renewcommand{\arraystretch}{0.7}
\caption{\emph{Left}: effects of each action; \emph{Right}: comparison of search length with different search methods and search space design.}
\vspace{-2mm}
\begin{subtable}[t]{0.4\linewidth}
\centering
\begin{tabular}{@{}cc|cc@{}}
\toprule
\textsc{In} & \textsc{Out} & Att & Spa \\
\midrule
\checkmark &            & 89.6 & 73.7 \\
           & \checkmark & 87.8 & 72.4 \\
\checkmark & \checkmark & \textbf{93.0} & \textbf{86.8} \\
\bottomrule
\end{tabular}
\label{tab:6a}
\end{subtable}
\hfill
\begin{subtable}[t]{0.56\linewidth}
\centering
\begin{tabular}{@{}l|ccc@{}}
\toprule
       & State & Length ($\downarrow$) & Budget \\
\midrule
MCTS   & 7     & 2.24   & 4 \\
A*     & 7     & 3.07   & 4 \\
\textbf{Ours}   & 4     & \textbf{1.87}   & 4 \\
\bottomrule
\end{tabular}
\label{tab:6b}
\end{subtable}
\label{tab:6}
\vspace{-2mm}
\end{table}

\subsection{In-depth Analysis}
\label{sec:analysis}
\begin{figure*}[t]
	\centering
	\includegraphics[width=1\linewidth]{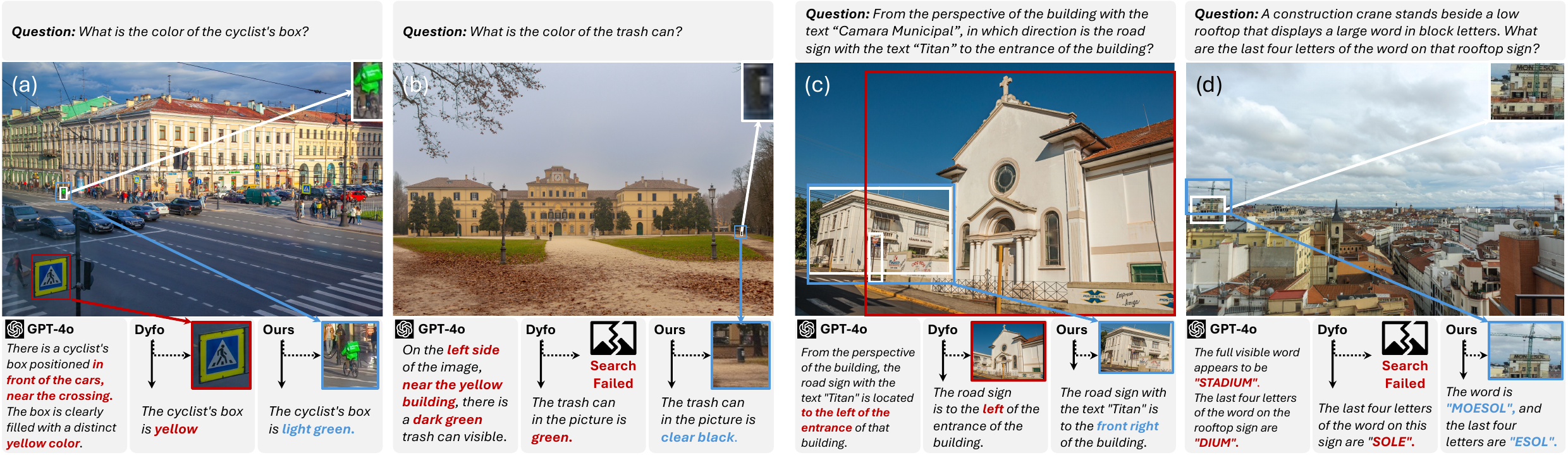}
    % \vspace{2mm}
	\caption{Case study of grounding results and model responses of GPT-4o, DyFo, and the proposed DeepScan. Panels (a–b) are from V* Bench; panels (c–d) are from TreeBench. Correct and incorrect responses are highlighted in \textcolor{local}{\textbf{blue}} and \textcolor{myred}{\textbf{red}} text, respectively. GT, evidence searched by Dyfo, and evidence localized by DeepScan are marked in the image using \emph{white}, \emph{red}, and \emph{blue} bboxes, respectively.}
	\label{fig:10}
    \vspace{-1mm}
\end{figure*}

\noindent\textbf{Case Study.}
\cref{fig:10} presents case studies from V* Bench and TreeBench. Specifically, GPT-4o and DyFo easily suffer from attention drift, leading to mislocalization and incorrect answers; in contrast, DeepScan accurately localizes evidence and produces interpretable answers (\cref{fig:10}a, c). Furthermore, DeepScan precisely grounds ultra-subtle evidence (area ratio ${<}1\permil$) and generates correct answers, demonstrating the effectiveness of Hierarchical Scanning (\cref{fig:10}b, d). TreeBench typically involves long queries and complex visual reasoning. Despite this, DeepScan consistently identifies evidence and enhances LVLMs via visually grounded reasoning, revealing its robustness (\cref{fig:10}c, d).

\begin{figure}[t]
	\centering
	\includegraphics[width=1\linewidth]{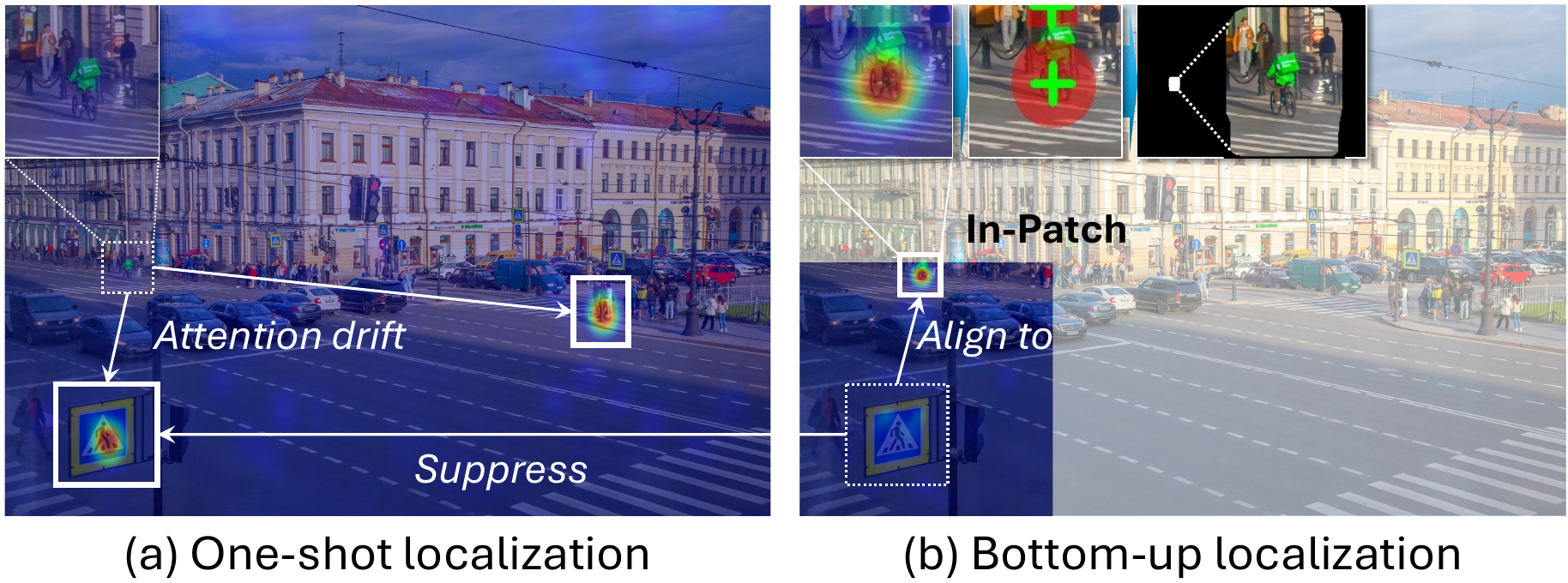}
	\caption{Qualitative analysis of the grounding paradigms with the attention map $S$ of the \emph{search expert} in~\cref{fig:10} (a).
    }
	\label{fig:11}
    \vspace{-2mm}
\end{figure}

\noindent\textbf{Analysis of Grounding Paradigms.}
The key insight of DeepScan is the \emph{bottom-up} grounding paradigm. To further reveal its fundamental superiority, we design a \emph{one-shot} variant of Hierarchical Scanning that performs \emph{image-level} rather than \emph{patch-wise} cue exploration, and provide qualitative and quantitative comparisons between the two variants.

\noindent\emph{$\diamondsuit$ Qualitative Analysis.} As shown in~\cref{fig:11}, the one-shot variant exhibits errors similar to DyFo with distinct experts (GroundingDINO \vs BLIP), where attention drifts toward the same object, ``box sign.'' This highlights the general impact of attention sink/drift on visual grounding. By contrast, the bottom-up variant suppresses the distractions and aligns attention with the correct target ``cyclist's box'', revealing the fundamental advantage of the bottom-up paradigm.

\begin{table}[t]
    \centering\footnotesize
    \setlength{\tabcolsep}{4.5pt}
    \renewcommand{\arraystretch}{1}
    \caption{Quantitative analysis of the grounding paradigms on V*.}
    \vspace{-2mm}
    \begin{tabular}{l | cccc}
    \toprule
    & Overall & Attribute & Spatial & Time ($\downarrow$)\\
    \midrule
    One-shot Localization  & 83.8 & 83.5 & 84.2 & 20.4s\\
    \textbf{Bottom-up Localization} & \textbf{90.6} & \textbf{93.0} & \textbf{86.8} & 24.5s\\
    \bottomrule
    \end{tabular}
    \label{tab:7}
    \vspace{-2.mm}
\end{table}

\noindent\emph{$\diamondsuit$ Quantitative Analysis.} As reported in~\cref{tab:7}, bottom-up localization clearly outperforms one-shot localization, particularly on fine-grained tasks. Moreover, the one-shot variant does not achieve a markedly higher inference speed than the bottom-up one. We attribute this to the proxy filtering in the evidence extraction, which keeps the LVLM judgment counts roughly comparable between the two paradigms.

\begin{figure}[t]
	\centering
	\includegraphics[width=0.94\linewidth]{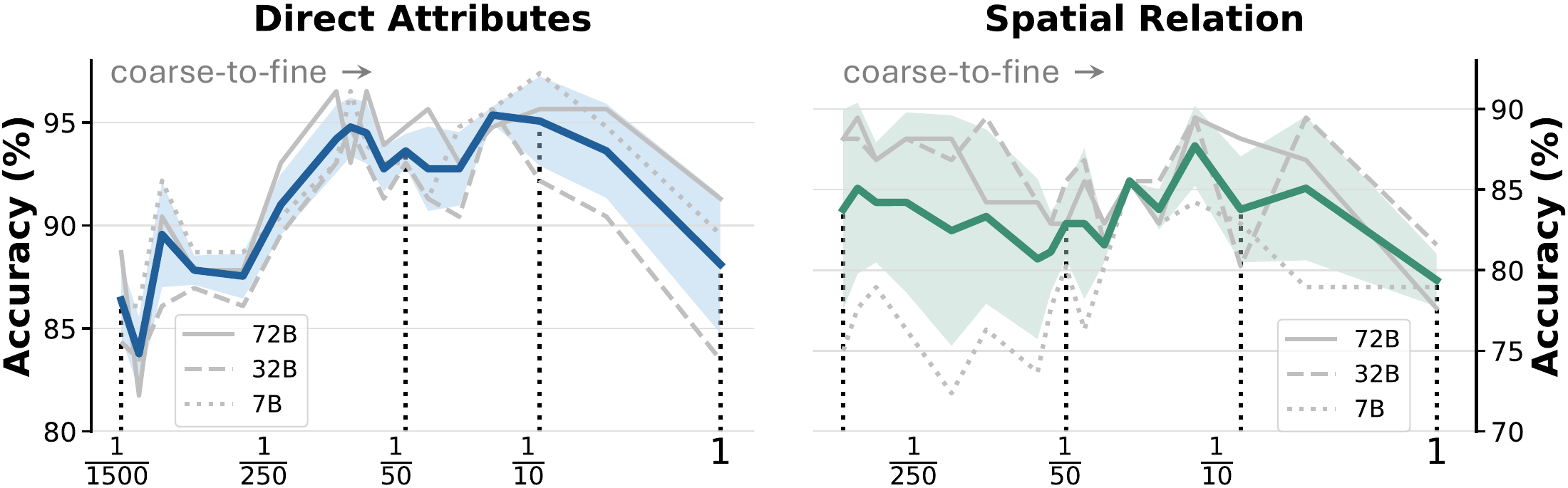}
	\caption{Performance of Qwen2.5-VL series with the varying grounding precision (measured by IoU) on V* Bench. 
    }
	\label{fig:12}
    \vspace{-3mm}
\end{figure}

\noindent\textbf{Impact of Grounding on LVLM Reasoning.} 
We study the reasoning performance in Qwen2.5-VL series under varying grounding precision, achieved by cropping the target with different sizes of surrounding context from the images on V*. As shown in~\cref{fig:12}, we point out two key insights:

\noindent\emph{$\diamondsuit$ More aggressive grounding is not always better.} Excessive zoom-in, \eg, IoU from $1/10$ to $1$, can remove essential context and thus degrade LVLM reasoning. Consistent with this observation, our selection policy avoids naively favoring smaller crops: the state reward~\cref{equ:14} pairs a size regularizer with LVLM feedback to discourage over-cropping.

\noindent\emph{$\diamondsuit$ Scaling effect benefits second-order reasoning more significantly than perception.} Under precise grounding, perception performance converges across model scales, while spatial reasoning gaps remain. 
This observation suggests using a lightweight LVLM for evidence judgment, and reserving a large-scale LVLM for Evidence-Enhanced Reasoning, preserving performance yet reducing latency.

\vspace{-0.5mm}
\section{Conclusion}
\vspace{-0.5mm}
We present DeepScan, a training-free framework for visually grounded reasoning in LVLMs via explicit evidence localization, recalibration, and integration before answering. Leveraging bottom-up Hierarchical Scanning, DeepScan mitigates effects of noisy context and precisely localizes critical visual content. It further refines the evidence view via Refocusing, yielding more accurate and grounded answers. Experiments show that DeepScan delivers superior performance across diverse visual tasks and substantially improves various LVLMs spanning architectures and parameter scales. Comprehensive ablations and analyses provide additional insights for the LVLM community.

\clearpage
\section*{Acknowledgments}
This work was supported by the National Natural Science Foundation of China (62176091) and the Natural Science Foundation of Chongqing (CSTB2024NSCQ-MSX0877). % and Shanghai AI Laboratory.

{
    \small
    \bibliographystyle{ieeenat_fullname}
    \bibliography{main}
}

\clearpage
\appendix                 
\setcounter{page}{1}      
\maketitlesupplementary

\section*{\textsc{Outline}}

\begin{itemize}[label={}, leftmargin=1.5em,itemsep=0.25em]
  \item \S~\ref{supp:discuss} \textbf{Discussions} includes the \emph{failure cases} of DeepScan, limitations and future work, and Broader Impacts, delivering valuable insights.
  
  \item \S~\ref{supp:detials} \textbf{Methodological Details} provides pseudocode, the exact prompts applied, and hyperparameter settings.
 
  \item \S~\ref{supp:comp} \textbf{Additional Results} conducts a comparison against strong baselines and reveals the state-of-the-art performance achieved by DeepScan on fine-grained tasks. 

  \item \S~\ref{supp:case} \textbf{Additional Cases} including the advantage of the \emph{bottom-up} grounding paradigm and the Example Model Outputs

\end{itemize}

\section{Discussions}
\label{supp:discuss}

\noindent\textbf{Failure Cases.}
\label{supp:errors}
Based on our analysis, DeepScan’s failures mainly fall into two categories.

\begin{figure}[t]
	\centering
	\includegraphics[width=0.71\linewidth]{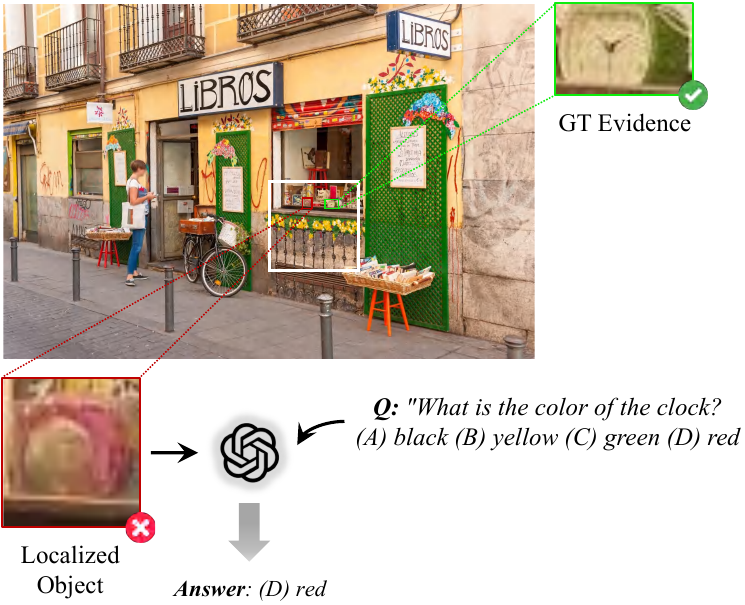}
	\caption{Case study of grounding failure} 
	\label{fig:supp1a}
\end{figure}

\noindent\emph{(1) Grounding failure.}
In direct-attribute recognition, when multiple visually similar objects remain within the evidence neighborhood, the experts may propose incorrect evidence and the LVLM may misjudge it, leading to an incorrect evidence localization; and thus the LVLM produces an incorrect answer based on wrong evidence (see~\cref{fig:supp1a}).

\begin{figure}[t]
	\centering
	\includegraphics[width=1\linewidth]{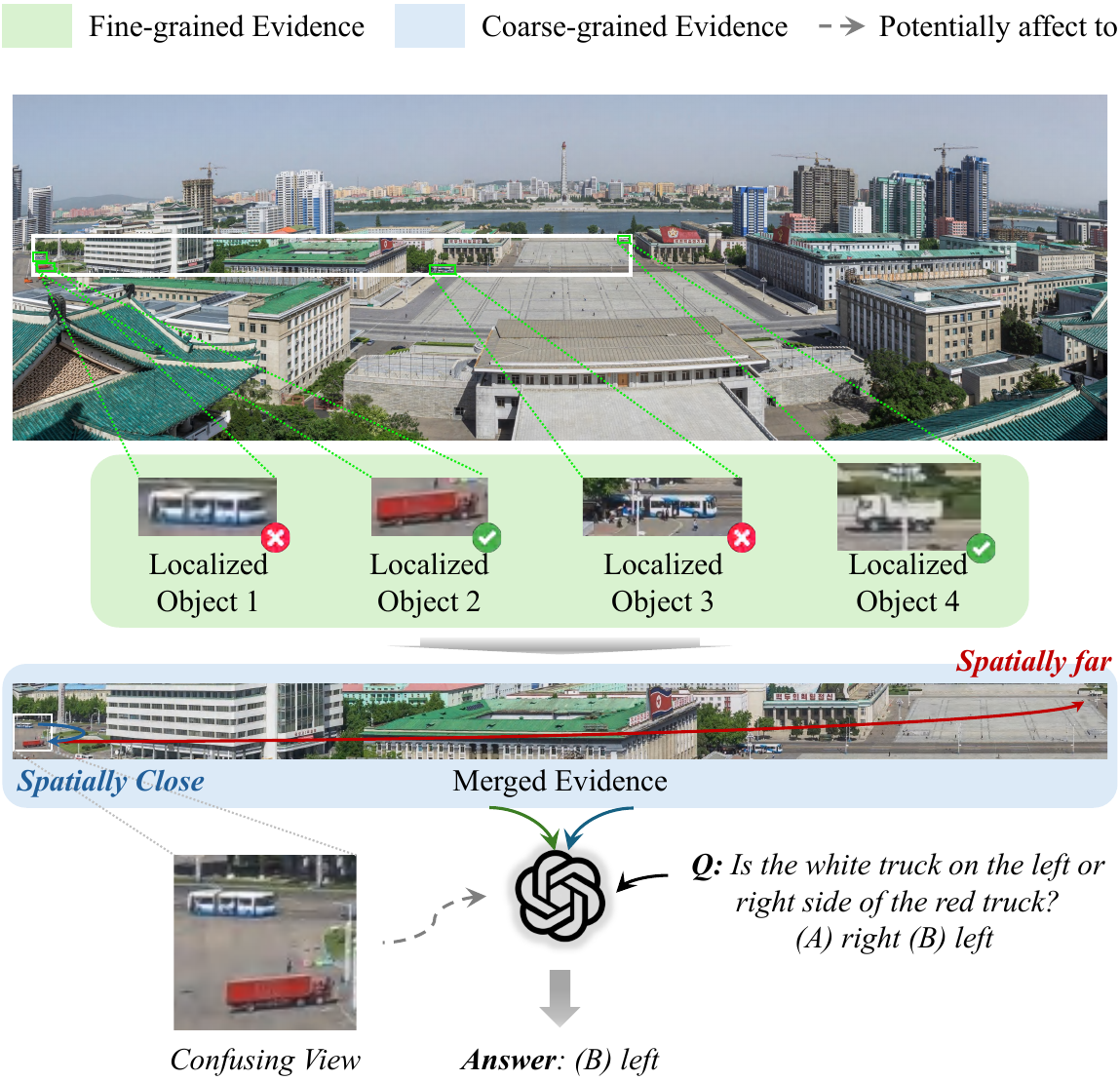}
	\caption{Case study of reasoning failure} 
	\label{fig:supp1b}
\end{figure}

\noindent\emph{(2) Reasoning failure.}
For spatial-relation reasoning with multiple pieces of evidence, DeepScan currently forms a merged evidence view via the minimal enclosing bounding box over all localized evidence. When the evidence is widely separated, this large crop introduces substantial inter-evidence noisy context that hampers reasoning; the issue is exacerbated when the merged view includes distracting content that conflicts with the correct answer (see~\cref{fig:supp1b}). This points to a promising direction: replace the minimal enclosing box with a generative composition that reassembles localized fine-grained evidence into a compact layout, suppressing inter-evidence context while preserving object attributes and spatial relations.


\noindent\textbf{Limitations.}
Compared with one-shot evidence-detection pipelines for visually grounded reasoning, DeepScan has higher inference latency. Nevertheless, as a test-time scaling paradigm, DeepScan offers a controllable performance-efficiency trade-off by tuning the patch size and the number of evidence proposals. In practice, its overall performance-efficiency profile remains acceptable relative to advanced visually grounded models (\eg, GPT-5), as detailed in ~\S\ref{supp:case} (Example Model Outputs). 
DeepScan's current implementation sets a relatively small patch size per image to capture fine-grained cues, which is unnecessary and inefficient for simple cases with salient evidence. In the future, we plan to develop an adaptive partition strategy that flexibly assigns a specific patch size within each grounding process based on stronger priors (\eg, evidence saliency), which is expected to further improve the performance-efficiency trade-off.

\noindent\textbf{Broader Impacts.}
We propose DeepScan, a training-free visually grounded reasoning framework that replaces brittle one-shot localization with hierarchical scanning and refocusing. By tying answers to concrete visual evidence and making contextual extent explicit, DeepScan can benefit domains needing reliable, fine-grained perception under clutter and occlusion. Examples include GUI agents (anchoring actions to the correct widget, explaining clicks), embodied manipulation (grasping small parts), and autonomous driving (partially occluded signs).
Potential risks include propagation of biases inherited from LVLMs and experts, and automation errors in safety-critical settings. Besides, as a test-time scaling paradigm, DeepScan introduces extra inference overhead, which may limit deployment in compute-constrained scenarios such as edge or mobile devices.

\begin{purpleprompt}[floatplacement=!t]{System Prompt}
You are an advanced image understanding assistant. You will be given an image and a question about it.
\end{purpleprompt}
\begin{purpleprompt}[floatplacement=!t]{User Prompt 1 (Evidence Decomposition)}
Task: List objects mentioned in text in List format. \\
Input text: \{question\} \\
Action: What objects are mentioned in original \\ text? List separated by commas. For example, from ``person with white trousers on the left or right side of the person in blue'', output ``[``person with white trousers'', ``person in blue'']''.
\end{purpleprompt}
\begin{purpleprompt}[floatplacement=!t]{User Prompt 2 (Evidence Judgment)}
I will provide you an image and a **question**: \\ \{question\}, please firstly determine whether the image contains the clues for answering the question or not (answer with **Yes** or **No**); then give the evidence of your decision.
\end{purpleprompt}
\begin{purpleprompt}[floatplacement=!t]{User Prompt 3 (View Completeness Justification)}
Question: Does the image fully contain every object in the list \{target\_\,list\}? Please treat ``fully contain'' as entirely within the frame (not truncated by image boundaries). Please firstly answer the question with **Yes** or **No**; then give the evidence of your decision. For example, if yes, list the evidence of each object (\eg, object: bbox [x1, y1, x2, y2] or a clear region description); if no, list the missing objects by name.
\end{purpleprompt}

\begin{algorithm}[t]
  \caption{Hierarchical Scanning \emph{with} Acceleration}
  \label{alg:1}
  \begin{algorithmic}[1]
    \Require image $I {\in} \mathbb{R}^{H \times W \times 3}$, question $q$, patch size $\{l_s, l_c\}$, candidate count $k$, kernel $\mathcal{K}$, $\mathcal{S}_r$, threshold $\tau_{\mathrm{area}}$, $\theta_{\mathrm{IoU}}$
    \Ensure Evidence set $\mathcal{E}=\{(b_i,e_i)\}$
    \State $\mathcal{E}_{\rm all}\!\gets\!\emptyset,\quad \mathcal{E}\!\gets\!\emptyset$
    \State $l \gets \textsc{SelectByLen}(\textsc{Lvlm}(\texttt{Prompt1}(q)), \{l_s,l_c\})$

    \For{\textbf{each} patch $p$ in $\textsc{Partition}(I,l)$}
        \State $S_p \gets \textsc{Search}(p,q)$ 
        \State $S_p^+ \gets \mathbb{I}\big(S_p \ge \textsc{OTSU}(S_p)\big)$ 
        \For{\textbf{each} $G \in \textsc{ConnComp}(S_p^+)$ with $|G|\!\ge\!\tau$}
            \State $d_G(i,j)\!\gets\!\inf_{\gamma\in\partial G}\| (i,j)-\gamma\|_2,\ \forall (i,j)\!\in\!G$
            \State $\tilde{S}_p \gets \textsc{Norm}(S_p)$, \quad $\tilde d\!\gets\!\textsc{Norm}(d_G)$
            \State $c^\star \gets \arg\max_{c\in G}\ \tilde S_p(c)\cdot \tilde d(c)$ 
            \State $c'\gets\textsc{LiftToImage}(c^\star,p\rightarrow I),\quad \mathcal{C}_p' \cup \{c'\}$
        \EndFor
    \EndFor

    \While{$\mathcal{C}_p'$ not empty}
        \State $c \gets \Call{Pop}{\mathcal{C}_p}$,\quad $m \gets \textsc{Segment}(I,c)$
        \State $m^+ \gets (\,m \bullet \mathcal{K}\,) \ \oplus\ \mathcal{S}_r$ 
        \State $b \gets \textsc{BBox}(m^+)$,\quad $e \gets \textsc{Crop}(I,b)$
        \If{$\mathrm{IoU}(b,b_i) \le \theta_{\rm IoU}$ for all $(b_i, e_i)\!\in\!\mathcal{E}_{\rm all}$}
            \State $\mathcal{E}_{\rm all}\!\gets\!\mathcal{E}_{\rm all}\cup\{(b,e)\}$
        \EndIf
        \State $I \gets I \odot (\mathbf{1}-m^+)$,\ \ $\mathcal{C}_p'\gets \{c'\in\mathcal{C}_p'\mid m^+(c')=0\}$ 
    \EndWhile

    \State $\mathcal{E}_k \gets \textsc{TakeKSmallestByArea}(\mathcal{E}_{\rm all},k)$ 
    \For{\textbf{each} $(b,e)\in \mathcal{E}_k$}
        \If{$\textsc{Lvlm}(e,\texttt{Prompt2}(q))=\texttt{Yes}$}
           \State $\mathcal{E}\gets \mathcal{E}\cup\{(b,e)\}$
        \EndIf
    \EndFor
    \State \Return $\mathcal{E}$
  \end{algorithmic}
\end{algorithm}
\begin{algorithm}[t]
  \caption{Refocusing}
  \label{alg:2}
  \begin{algorithmic}[1]
    \Require image $I\!\in\!\mathbb{R}^{H\times W\times3}$, question $q$, target list $t$, evidence set $\mathcal{E}$
    \Ensure refined view $V$
    \State $\mathcal{R}\gets \emptyset,\quad b_{\rm m}\gets \bigcup_{(b,e)\in\mathcal{E}}b$ %\Comment{Minimal enclosing box for all evidence content}
    \State $e_{\rm m}\gets \Call{Crop}{I,\, b_{\rm m}},\quad V_1\gets e_m$
    \State $V_2\gets \Call{In}{V_1,q},\quad V_3\gets \Call{Out}{V_1},\quad V_4\gets \Call{In}{V_3,q}$
    \For{$V\in \{V_1, V_2, V_3, V_4\}$}
    \If{$\Call{Lvlm}{V,\texttt{UserPrompt3}(t)}=\texttt{Yes}$} %\Comment{Complete view}
    \State $h,w\gets \Call{Shape}{V}$
    \State $R\gets HW/hw$
    \Else
    \State $R\gets 0$
    \EndIf
    \State $\mathcal{R}\gets\mathcal{R}\,\cup\,\{R \}$
    \EndFor
    \State $i^*\gets\argmax_{i\in\{1,2,3,4\}} \mathcal{R}(i)  $
    \State \Return $V_{i^*}$
  \end{algorithmic}
\end{algorithm}

% \clearpage
\section{Methodological Details}
\label{supp:detials}

\noindent\textbf{Prompts.} We provide below the prompts involved in~\cref{Sec:3}: (i) \emph{Evidence Decomposition} (to set the patch size), (ii) \emph{Evidence Judgment} in \emph{Hierarchical Scanning}, and (iii) \emph{View Completeness Justification} in \emph{Refocusing}.

\begin{table*}[t]
  \centering\footnotesize
  \setlength{\tabcolsep}{10pt}
  \renewcommand{\arraystretch}{1.1}
  \caption{Comparisons between the proposed DeepScan and existing visually grounded reasoning approaches. Like most test-time scaling paradigms, DeepScan is easier to scale up. Furthermore, DeepScan introduces hierarchical scanning and refocusing for a robust bottom-up evidence localization and recalibration and leverages hybrid granular evidence to enable LVLM to produce higher-quality answers.
  }
  \begin{tabular}{l *{7}{c}}
    \toprule
    \multicolumn{1}{l}{Methods} &
    \multicolumn{1}{c}{Venue} &
    \thead{Ease to\\Scaling} &
    \thead{External\\Experts} &
    \thead{Search\\Strategy} &
    \thead{Grounding\\Paradigm} &
    \thead{Evidence\\Granularity} &
    \thead{Inference\\Latency} \\
    \midrule
    Seal~\cite{wu2024vstar}                     & CVPR'24     & \xmark & \xmark & LLM-guided Search  & Coarse-to-Fine & Fine  & High \\
    Dyfo~\cite{li2025dyfo}                      & CVPR'25     & \cmark & \cmark & Detection + MCTS   & Coarse-to-Fine  & Coarse  & High \\
    DeepEyes~\cite{zheng2025deepeyes}           & NeurIPS'25  & \xmark & \xmark & Generative Bbox  & Coarse-to-Fine   & Fine   & High \\
    PixelReasoner~\cite{su2025pixelreasoner}    & NeurIPS'25  & \xmark & \xmark & Generative Bbox    & Coarse-to-Fine & Coarse   & High \\
    ViGoRL~\cite{sarch2025grounded}             & NeurIPS'25 & \xmark & \xmark & Generative Bbox  & Coarse-to-Fine   & Fine   & High \\
    ZoomRefine~\cite{yu2025zoom}                & NeurIPS'25  & \cmark & \xmark & Generative Bbox    & Coarse-to-Fine & Coarse    & Medium \\
    TreeVGR~\cite{wang2025traceable}            & Preprint, Jul & \xmark & \xmark & Generative Bbox    & Implicit Search  & NA  & Low \\
    Thyme-VL~\cite{zhang2025thyme}              & Preprint, Aug & \xmark & \xmark & Generative Code  & Coarse-to-Fine   & Coarse     & High \\
    \multirow{2}{*}{DeepScan (Ours)}   & \multirow{2}{*}{--}    & \multirow{2}{*}{\cmark} & \multirow{2}{*}{\cmark} & Hierarchical Scanning     & \multirow{2}{*}{Bottom-up} & \multirow{2}{*}{Hybrid} & \multirow{2}{*}{High} \\
    & & & & + Refocusing & & & \\
    \bottomrule
  \end{tabular}
  \label{supp:tab1}
\end{table*}

\noindent\textbf{Hyper-Parameters.} In \emph{Local Cue Exploration}, the area threshold for noisy-cue filtering is set to \(50\) pixels. In \emph{Multi-Scale Evidence Extraction}, morphological post-processing uses a \(5{\times}5\) flat structuring element \(\mathcal{K}\) and a disk \(\mathcal{S}_r\) with radius \(r{=}20\). The IoU threshold for filtering similar evidence is \(\theta_{\rm IoU}{=}0.3\). We set \(k{=}10\), \ie, retain only the 10 smallest pairs \((b,e)\in\mathcal{E}\) for acceleration. For \emph{Refocusing}, to prevent undersized crops, we pad the visual expert's detections (\(\mathcal{B}\)) by \(28\) pixels on all sides, and set the scaling factor to \(s{=}1.5\). For \emph{LVLM Querying}, the maximum output length is \(50\) for \emph{Evidence Decomposition}, \emph{Evidence Judgment}, and \emph{View Completeness Justification}, and \(1024\) for \emph{Evidence-Enhanced Reasoning}. During inference, we use temperature \(t{=}0\) with a fixed random seed (\(13\)). Beam search and top-\(k\) sampling are disabled by default.

\noindent\textbf{Pseudocodes.}
Algorithmic details for Hierarchical Scanning and Refocusing are provided in \cref{alg:1,alg:2}. Here, the prompts \texttt{Prompt1}, \texttt{Prompt2}, and \texttt{Prompt3} are specified in the \textbf{Prompt} subsection, \ie, ``User Prompt 1-3''; $\texttt{Prompt1}(x)$ denotes instantiating the prompt with the string $x$ via template-based substitution. In addition, \textsc{LiftToImage} maps patch coordinates to the full-image coordinate system. \textsc{Close} and \textsc{Dilate} indicate morphological closing and dilation, respectively. The operator $\odot$ denotes element-wise multiplication.

\begin{table}[t]
    \centering%\small
    \setlength{\tabcolsep}{5pt}
    \renewcommand{\arraystretch}{1}
    \caption{Additional comparison results on V* Bench between our method with existing
    baselines. We produced the results$^\dagger$ with the provided official codes for fair comparisons.}
    \begin{tabular}{l | ccc}
    \toprule
    & Overall & Attribute & Spatial\\
    \midrule
    o3               &95.0 & - & - \\  
    \midrule
    Seal             &74.8  &76.3 &75.4 \\
    Dyfo-L           &62.7  &53.9 &59.2  \\
    Dyfo-Q           &80.0  &82.9 &81.2 \\
    \midrule
    Qwen2.5-VL-7B    &74.3    &77.4   &69.7 \\
    PixelReasoner    &80.6    &83.5   &76.3 \\
    
    Thyme-VL         &82.2    &83.5   &80.3 \\
    ZoomRefine$^\dagger$      &82.2   &85.3 &77.6 \\
    Dyfo$^\dagger$   &84.3    &82.6   &86.8        \\
    TreeVGR$^\dagger$&85.9    & 86.1  &85.5  \\
    ViGoRL           &86.4    &-      &-  \\
    DeepEyes         &90.0    &92.1   &86.8 \\
    \textbf{DeepScan} ($k=10$) &90.6    &93.0   &86.8  \\
    \textbf{DeepScan} ($k=\infty$) &91.1    &93.9   &86.8  \\
    \midrule
    Qwen2.5-VL-72B   &84.8    &90.8   &80.9   \\
    \textbf{DeepScan-72B} ($k=10$)     &93.7   &93.9   &93.4 \\
    \textbf{DeepScan-72B} ($k=\infty$) &94.2   &94.8   &93.4 \\
    \bottomrule
    \end{tabular}
    \label{supp:tab2}
\end{table}

 \begin{figure*}[t]
	\centering
	\includegraphics[width=0.92\linewidth]{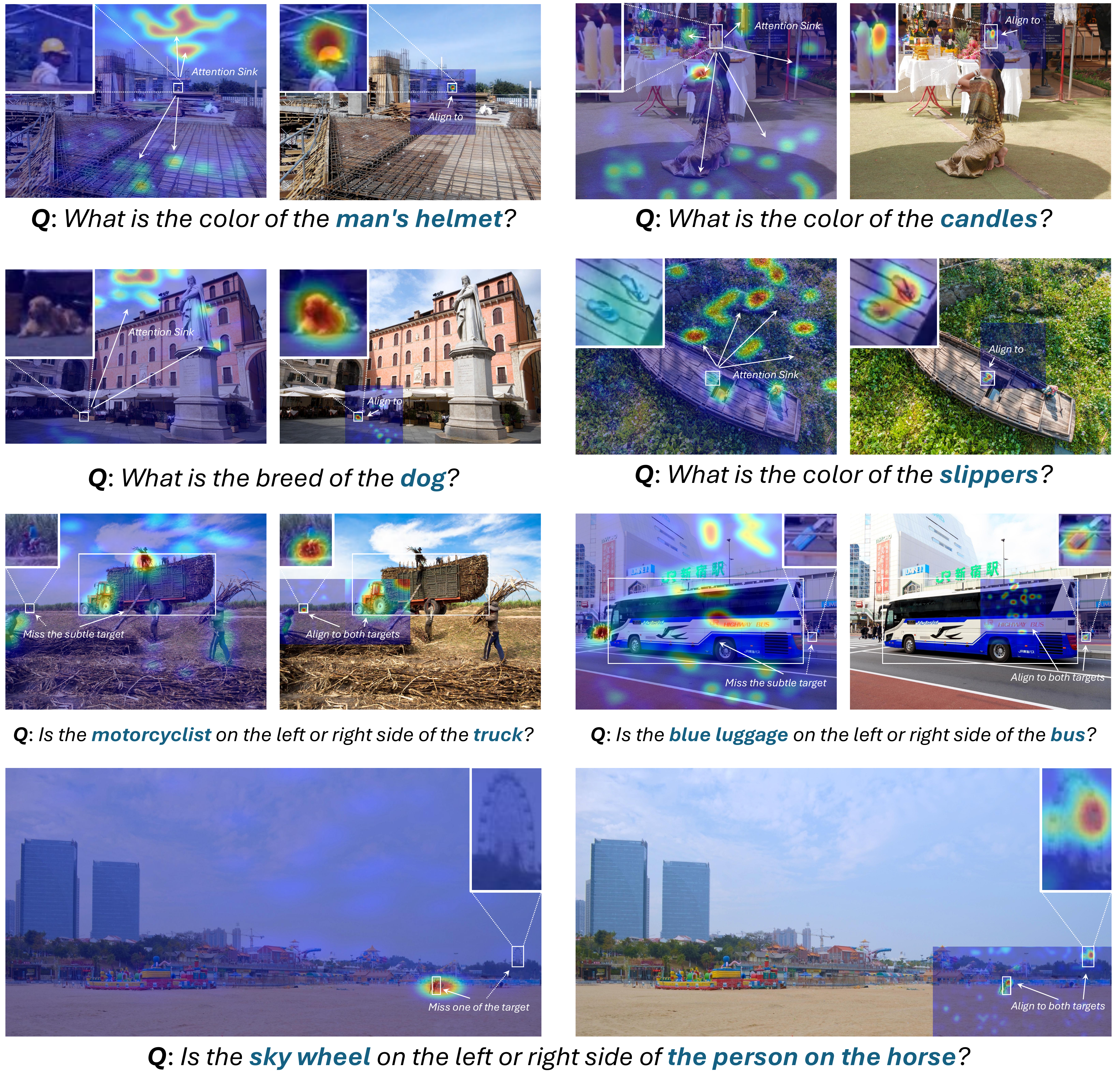}
    \vspace{-5mm}
	\caption{Qualitative analysis of the grounding paradigms with the attention map $S$ of the \emph{search expert} 
    } 
	\label{fig:supp2}
\end{figure*}

\section{Additional Results}
\label{supp:comp}

\noindent\textbf{Baselines.} 
SEAL~\cite{wu2024vstar}, PixelReasoner~\cite{su2025pixelreasoner}, TreeVGR~\cite{wang2025traceable}, DeepEyes~\cite{zheng2025deepeyes}, ViGoRL~\cite{sarch2025grounded}, and Thyme-VL~\cite{zhang2025thyme} depend on learned localization modules or RL-based decision controllers for search. Upgrading these methods to a stronger LVLM typically requires retraining these components. In contrast, DyFo~\cite{li2025dyfo} performs MCTS over focus actions with a visual expert, and ZoomRefine~\cite{yu2025zoom} provides an unlearning variant that leverages LVLM prior for visually grounded reasoning through prompt engineering; both scale more readily across larger LVLM backbones. Our method is easier to scale up and introduces a novel bottom-up hierarchical scanning and refocusing for context-optimal views, complemented by a hybrid evidence memory, yielding better robustness than existing approaches.

\noindent\textbf{Performance Comparison.}
\label{supp:result}
As shown in~\Cref{supp:tab2}, \textbf{DeepScan} attains \textbf{90.6}\% at $k{=}10$ and \textbf{91.1}\% at $k{=}\infty$ based on Qwen2.5-VL-7B, which substantially outperforms both the training-free ZoomRefine$^\dagger$ (82.2\%) and DyFo$^\dagger$ (84.3\%) and even surpasses the RL-based DeepEyes (90.0\%), ViGoRL (86.4\%), TreeVGR (85.9\%), Thyme-VL (82.2\%), and PixelReasoner (80.6\%) on the V* Benchmark. Specifically, on the \emph{Attribute} subset, DeepScan reaches 93.0-93.9\% while maintaining \emph{Spatial} at 86.8\%, indicating that bottom-up recovery plus refocusing preserves necessary context for reasoning. When scaling the LVLM backbone to 72B, \textbf{DeepScan-72B} achieves \textbf{94.2}\% overall and 93.4\% on the \emph{Spatial} subset, a 9.4\% gain over the Qwen2.5-VL-72B (84.8\%). These results demonstrate DeepScan as a training-free framework that consistently improves diverse LVLMs and benefits from model scaling, with a controllable performance-latency trade-off via top-$k$ selection.

\begin{table*}[t]
        \centering
		% \footnotesize
		\renewcommand\arraystretch{1}
		\setlength{\tabcolsep}{1.5mm}{
			\begin{tabular}{l | c | c c c | c c c }
				\toprule	
				& Qwen2.5-VL 7B & DeepEyes & Zoom Refine & Dyfo & Hierarchical Scan & Refocus & DeepScan \\
				\midrule      
                Accuracy (\%)     & 75.4    & 89.0  & 82.7  & 83.8     & 84.8  & 89.5   & 90.1 \\
                Token Cost (k)    & 3.5     & 13    & 5.1   & 7.2      & 6.5   & 8.1    & 8.4 \\
                \emph{End-to-End} Latency (s)       & 0.4     & 6.9  & 0.9      &  2.4 &  2.2   &  3.0  & 3.1 \\
				\bottomrule		
			\end{tabular}}
            % \vspace{-2.5mm}
            \caption{Performance--efficiency comparison on V* Benchmark. \emph{End-to-End} Latency is the wall-clock time \emph{per sample} from input to final output, \emph{including all auxiliary steps} (\eg, Evidence Decompositions, Expert Calls, Post-Processing, Evidence Judgments).}
            \label{app-tab:1}
\end{table*}

\noindent\textbf{Engineering Optimization.} Our standard implementation of DeepScan relies on nested loops using the \emph{Hugging face transformers backend}  (as detailed in~\cref{alg:1,alg:2}). However, this implementation is severely bottlenecked by inefficient \textbf{sequential computation} during the Scanning and Refocusing stages. Furthermore, the frequent communication overhead between the visual expert and the LVLM leads to poor GPU utilization. Compounded by the \emph{lack of modern primitives} in the standard \textbf{HF Transformers backend}, these factors result in substantial inference latency (as shown in~\cref{fig:8}). To address this, we introduce a suite of engineering optimizations that exploit DeepScan's inherent algorithmic properties, significantly enhancing its viability for latency-sensitive applications.
Specifically, unlike MCTS-based methods (\eg, Dyfo), DeepScan is built on \emph{deterministic sampling}, enabling it to fully benefit from parallel acceleration via batching. Hence, we implemented batch processing for \textbf{(a)} attention map calculation, \textbf{(b)} top-$k$ candidate judgment, and \textbf{(c)} evidence view justification, and further parallelized post-processing routines to reduce tail latency. This strategy compresses many sequential interactions into a \emph{single batched} search expert call and \emph{three} LVLM forward passes \emph{with extended context}, thus \emph{substantially improving GPU utilization and reducing GPU idle time}. Moreover, we migrated to the \emph{vLLM backend}, leveraging PagedAttention and optimized CUDA kernels. \emph{Collectively, these standard engineering optimizations yield an \textbf{$\sim8\times$ speedup} over our previous implementation.}

\noindent\textbf{Performance-Latency Trade-off.} Integrating the above optimizations, we re-evaluated DeepScan via \emph{VLMEvalKit} (with vLLM backend) on 4$\times$L20 GPUs. As shown in~\Cref{app-tab:1}, compared to \emph{DeepEyes} that relies on multi-turn tool executions, DeepScan exhibits striking superiority. Specifically, DeepEyes suffers a severe 6.9s latency and 13k token cost, while DeepScan attains higher accuracy (90.1\% \emph{vs.} 89.0\%) with $\sim$2.2$\times$ faster speed (3.1s) as well as $\sim$35\% fewer tokens. This reveals that batched deterministic sampling effectively bypasses the sequential overhead of agentic paradigms, yielding superior reasoning with higher efficiency. Against the MCTS-based method \emph{Dyfo}, DeepScan also offers a favorable trade-off. It incurs a marginal latency overhead (3.1s \emph{vs.} 2.4s) but delivers a substantial $+6.3\%$ accuracy gain. Notably, our \emph{Hierarchical Scan} alone outperforms Dyfo (84.8\% \emph{vs.} 83.8\%) with lower latency (2.2s). 

In fine-grained scenarios, the target occupies a minuscule fraction of the image, resulting in an inherently low signal-to-noise ratio (SNR) where the effective visual signal is overwhelmed by context noise. In these cases, search-based methods struggle to identify a reliable initial anchor, frequently causing exhaustive tree expansion to degenerate into an inefficient random walk. By slicing the image and evaluating patches via a single batched forward pass, our hierarchical scan explicitly isolates the target, drastically amplifying the local SNR. Coupled with the subsequent \emph{Refocus}---which requires merely 0.8s but yields a massive $+4.7\%$ performance leap---these results reveal a profound insight: systematically scaling compute to deterministically maximize visual SNR is fundamentally more efficient than heuristic search-tree expansion.

\section{Additional Visualization}
\label{supp:case}

\subsection{Qualitative Analysis of Grounding Paradigm}
\emph{Direct Attributes.} For attribute questions (\eg, helmet/slippers/candles color, dog breed), the \emph{one-shot} variant tends to lock onto globally salient but irrelevant regions, a typical \emph{attention sink/drift} failure, yielding mis-localization or missing the true fine-grained cue. By contrast, our \emph{bottom-up} paradigm begins with local cue exploration and iteratively recenters the view on the true evidence, effectively “pulling” attention back to the correct object—consistent with the figure’s transitions from \emph{Attention Sink} to \emph{Align to} on attribute examples. This behavior aligns with the paper’s analysis that top-down, one-shot localization is fragile under noisy context, whereas bottom-up scanning suppresses distractions and aligns attention with the correct target.

\emph{Spatial Relations.} Relation questions (\eg, left/right of motorcyclist vs.\ truck, sky wheel vs.\ rider, blue luggage vs.\ bus) require jointly grounding \emph{two} targets. The one-shot variant often fixates on a single salient object and \emph{misses the subtle counterpart}, leading to incorrect relational judgments; bottom-up scanning progressively resolves and aligns to \emph{both} entities, producing reliable spatial decisions—exactly reflected by the figure’s captions “Miss the subtle target / Miss one of the target” versus “Align to both targets.” Moreover, after evidence is recovered, \emph{Refocusing} further calibrates the context window to an evidence-centric view, helping retain only the necessary surrounding context for reasoning.

\subsection{Example Model Outputs}

To illustrate the advantages of our approach, we present side-by-side qualitative examples comparing \textbf{DeepScan} with strong LVLMs—\mbox{GPT-5} (run in its advanced reasoning mode), \mbox{GPT-4o}, and \mbox{Qwen3-VL-235B-A22B}—as well as training-free baselines \emph{ZoomRefine} and \emph{DyFo}. For a fair comparison across training-free methods, ZoomRefine, DyFo, and \textbf{DeepScan} all use \mbox{Qwen2.5-VL-72B} as the LVLM backbone. Each example reports both the visual grounding and the model’s answer.
The following examples confirm that \textbf{DeepScan} delivers stronger visually grounded reasoning—superior evidence localization and more accurate answers—than the two representative training-free baselines. Moreover, \textbf{DeepScan} outperforms \mbox{GPT-4o} and the larger-scale \mbox{Qwen3-VL-235B-A22B}, despite using a smaller \mbox{Qwen2.5-VL-72B} backbone, in both perception and reasoning. Notably, its end-to-end inference latency is of the same order as \mbox{GPT-5}, indicating a favorable performance–efficiency trade-off.

\clearpage
\begin{figure*}[htbp]
	\centering
	\includegraphics[width=1\linewidth]{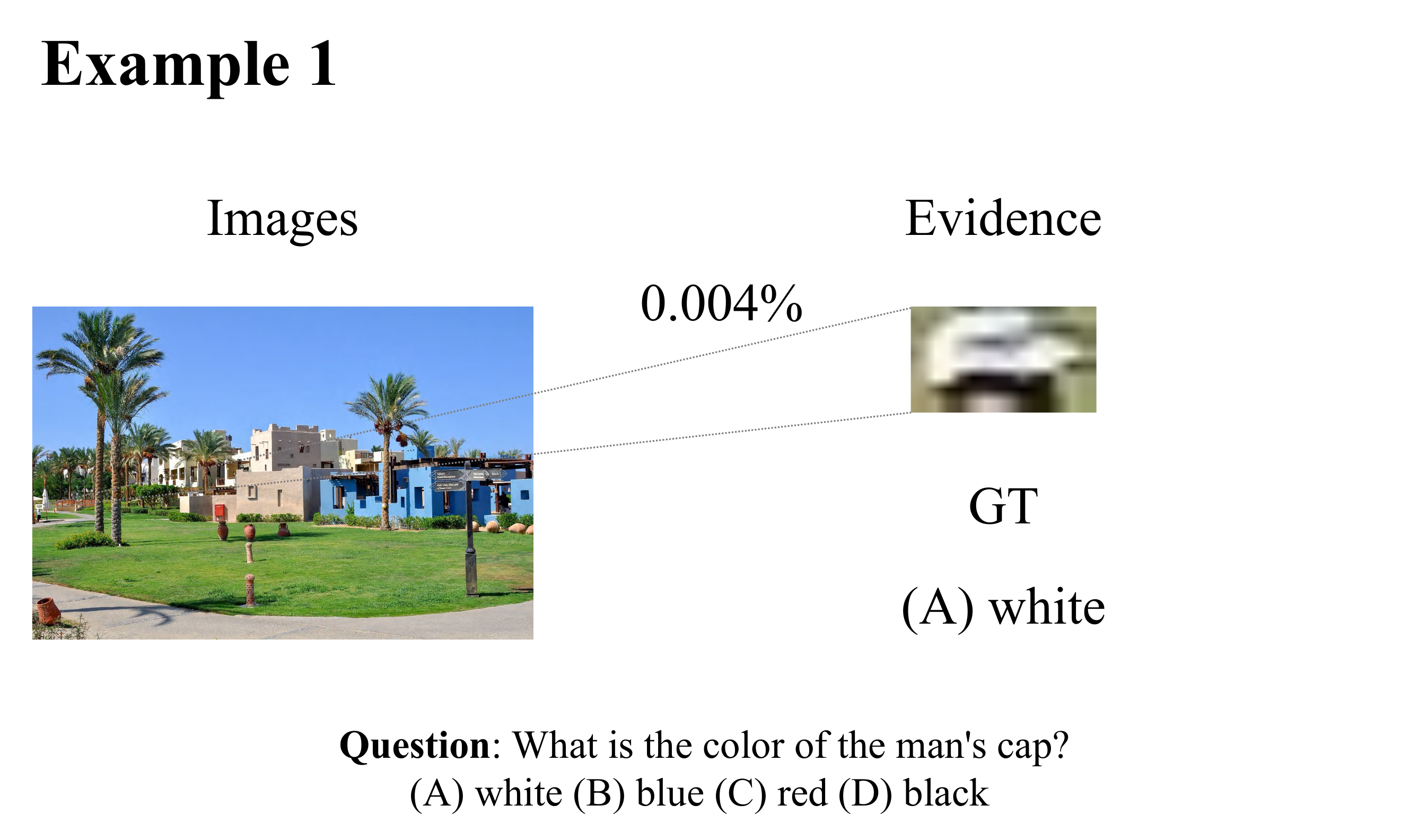}
\end{figure*}
\begin{figure*}[htbp]
	\centering
	\includegraphics[width=1\linewidth]{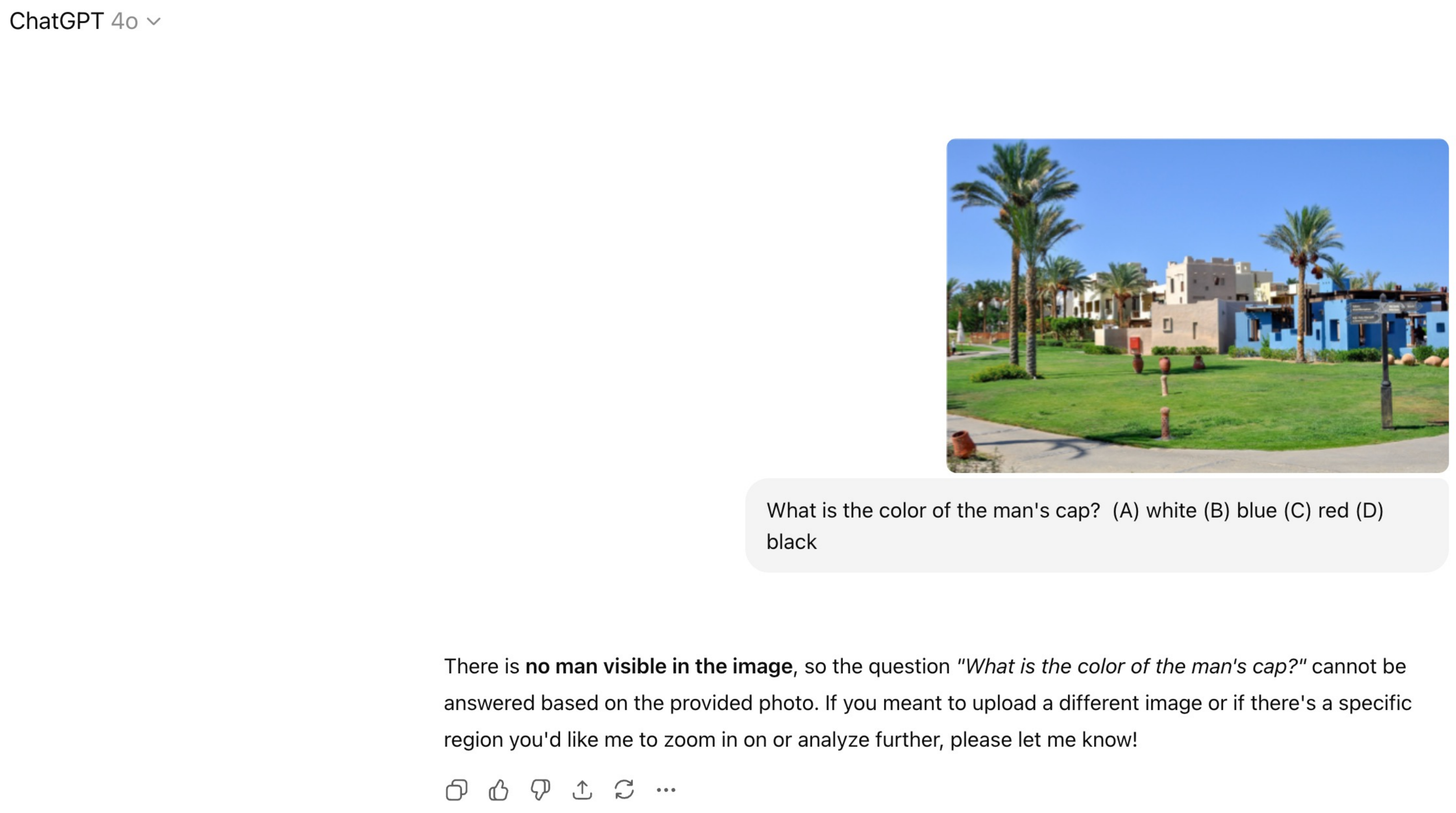}
\end{figure*}
% \begin{figure*}[t]
% 	\centering
% 	\includegraphics[width=1\linewidth]{fig/examples/exp1/gpt5-t.pdf}
% \end{figure*}
\begin{figure*}[htbp]
	\centering
	\includegraphics[width=1\linewidth]{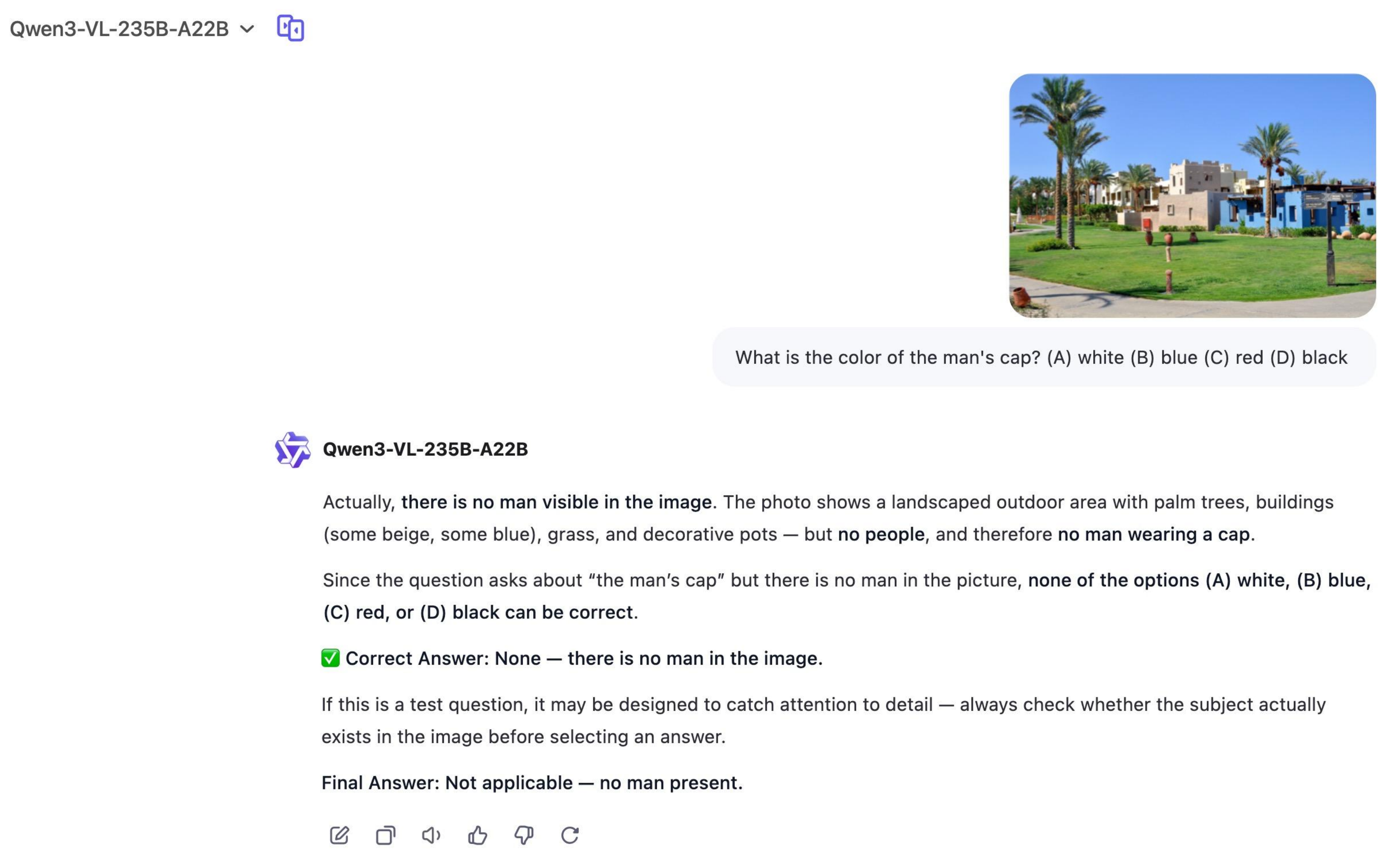}
\end{figure*}
\begin{figure*}[htbp]
	\centering
	\includegraphics[width=1\linewidth]{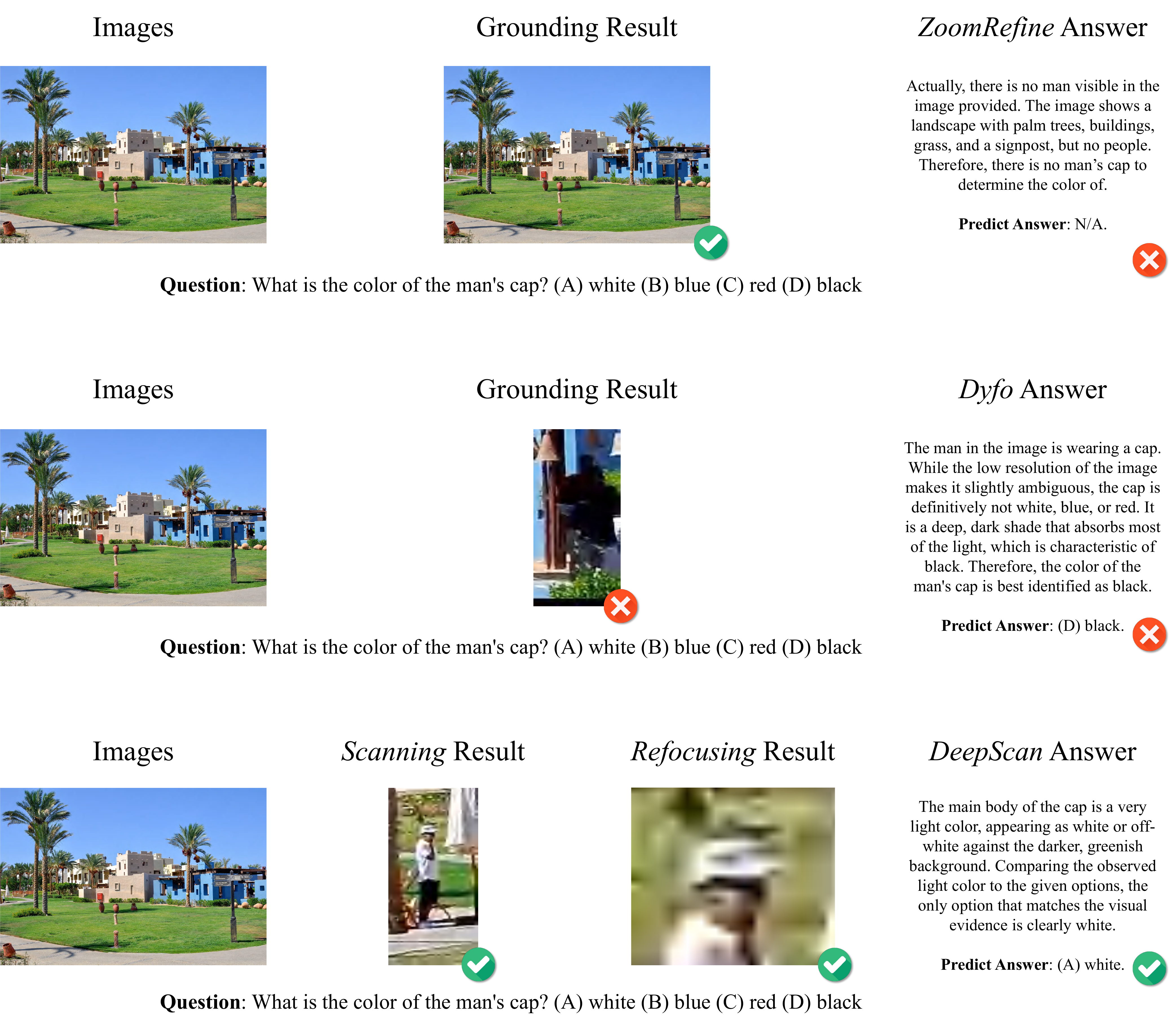} 
\end{figure*}

\end{document}